\documentclass[runningheads]{llncs}
\usepackage{graphicx}
\usepackage{amsmath,amssymb} 
\usepackage{color}
\usepackage{tikz}
\usepackage{subfigure}
\usepackage{multirow}
\usepackage{makecell}
\usepackage{enumerate}
\usepackage{url}
\usepackage{booktabs}
\usepackage{radarchart}
\usepackage{ulem}
\usepackage{subfig}

\makeatletter
\newcommand{\rmnum}[1]{\romannumeral #1}
\newcommand{\Rmnum}[1]{\expandafter\@slowromancap\romannumeral #1@}
\makeatother

\begin{document}
\pagestyle{headings}
\mainmatter

\def\ACCV20SubNumber{***}  

\title{DOLPHINS: Dataset for Collaborative Perception enabled Harmonious and Interconnected Self-driving} 
\titlerunning{DOLPHINS}
%
\author{Ruiqing Mao\inst{1} \and
Jingyu Guo\inst{1} \and
Yukuan Jia\inst{1} \and
Yuxuan Sun\inst{1} \and
Sheng Zhou\inst{1} \and
Zhisheng Niu\inst{1}}
\authorrunning{R. Mao et al.}
%
\institute{Department of Electronic Engineering, Tsinghua University, Beijing 100084, China\\
\email{\{mrq20,jyk20\}@mails.tsinghua.edu.cn}\\
\email{1078062607@qq.com}\\
\email{sunyuxuan@mail.tsinghua.edu.cn}\\
\email{\{sheng.zhou,niuzhs\}@tsinghua.edu.cn}}

\maketitle

\begin{abstract}
Vehicle-to-Everything (V2X) network has enabled collaborative perception in autonomous driving, which is a promising solution to the fundamental defect of stand-alone intelligence including blind zones and long-range perception. However, the lack of datasets has severely blocked the development of collaborative perception algorithms. In this work, we release DOLPHINS: Dataset for cOLlaborative Perception enabled Harmonious and INterconnected Self-driving, as a new simulated {\it large-scale various-scenario multi-view multi-modality} autonomous driving dataset, which provides a ground-breaking benchmark platform for interconnected autonomous driving. DOLPHINS outperforms current datasets in six dimensions: temporally-aligned images and point clouds from both vehicles and Road Side Units (RSUs) enabling both {\it Vehicle-to-Vehicle (V2V)} and {\it Vehicle-to-Infrastructure (V2I)} based collaborative perception; 6 typical scenarios with dynamic weather conditions make the most {\it various} interconnected autonomous driving dataset; meticulously selected viewpoints providing {\it full coverage} of the key areas and every object; 42376 frames and 292549 objects, as well as the corresponding 3D annotations, geo-positions, and calibrations, compose the {\it largest} dataset for collaborative perception; Full-HD images and 64-line LiDARs construct {\it high-resolution} data with sufficient details; well-organized APIs and open-source codes ensure the {\it extensibility} of DOLPHINS. We also construct a benchmark of 2D detection, 3D detection, and multi-view collaborative perception tasks on DOLPHINS. The experiment results show that the raw-level fusion scheme through V2X communication can help to improve the precision as well as to reduce the necessity of expensive LiDAR equipment on vehicles when RSUs exist, which may accelerate the popularity of interconnected self-driving vehicles. DOLPHINS is now available on \url{www.dolphins-dataset.net}.
\end{abstract}

\section{Introduction}

\begin{figure}
\centering
\subfigure[Stand-alone]{\includegraphics[width=4cm]{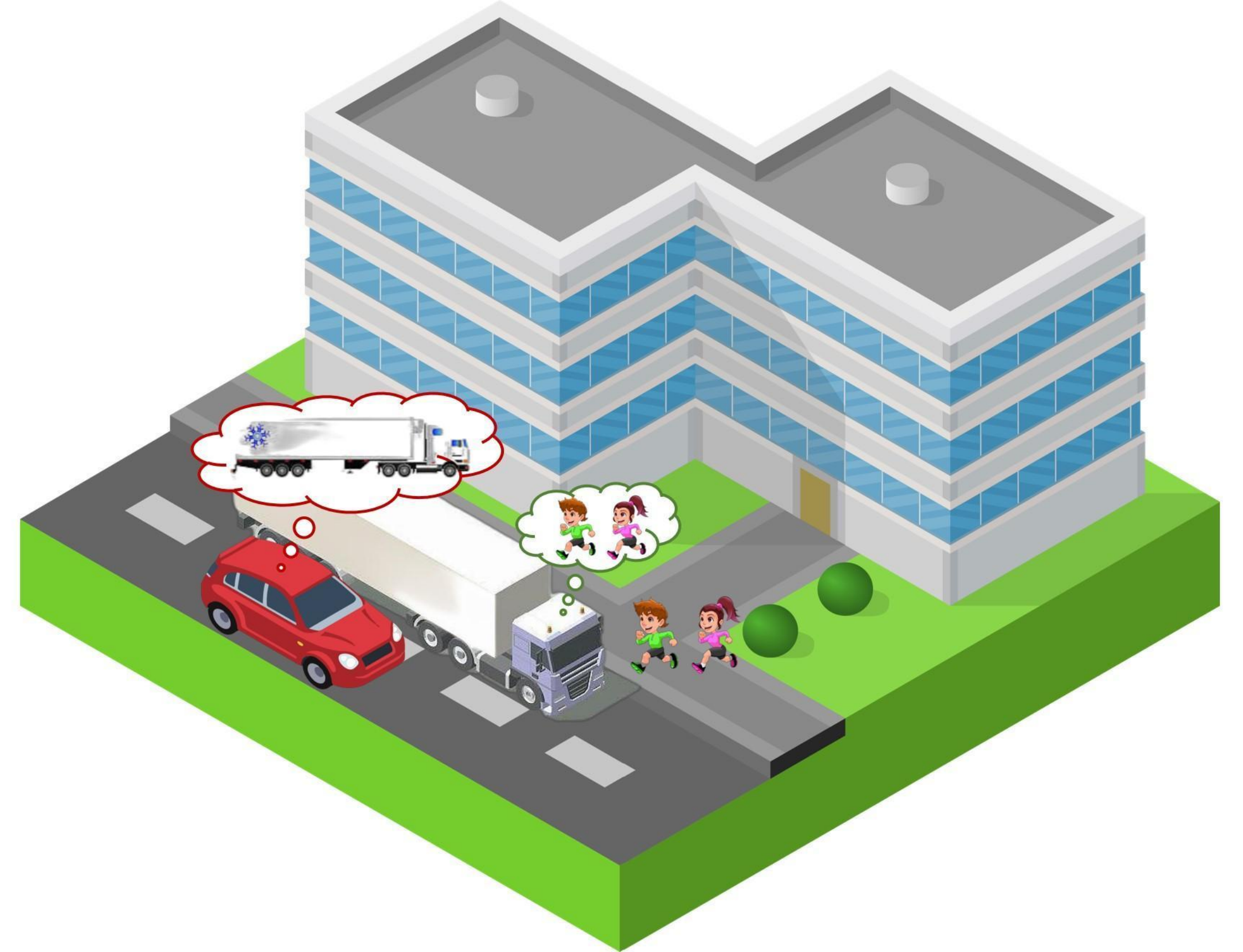}}
\subfigure[V2X communication]{\includegraphics[width=4cm]{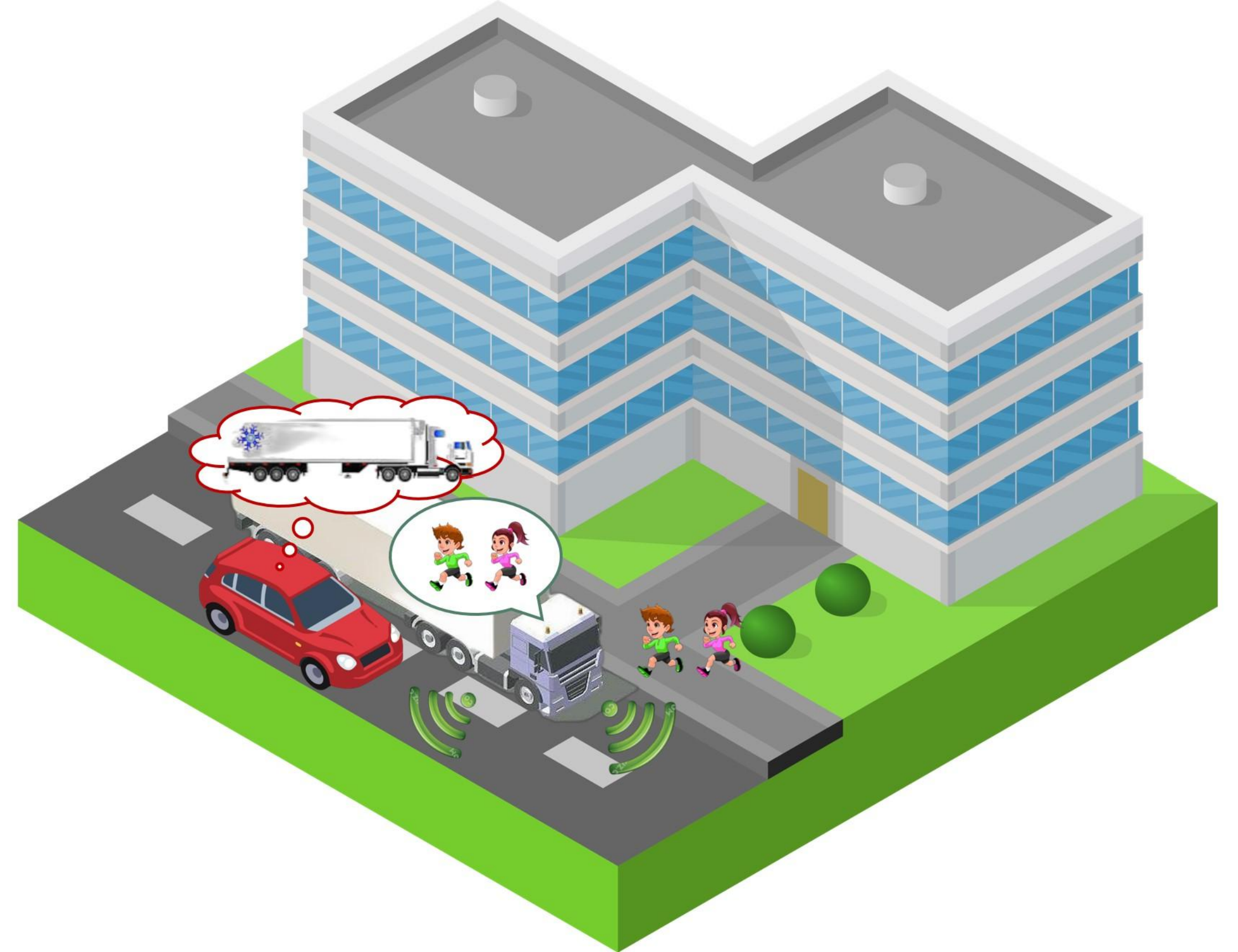}}
\subfigure[Collaborative perception]{\includegraphics[width=4cm]{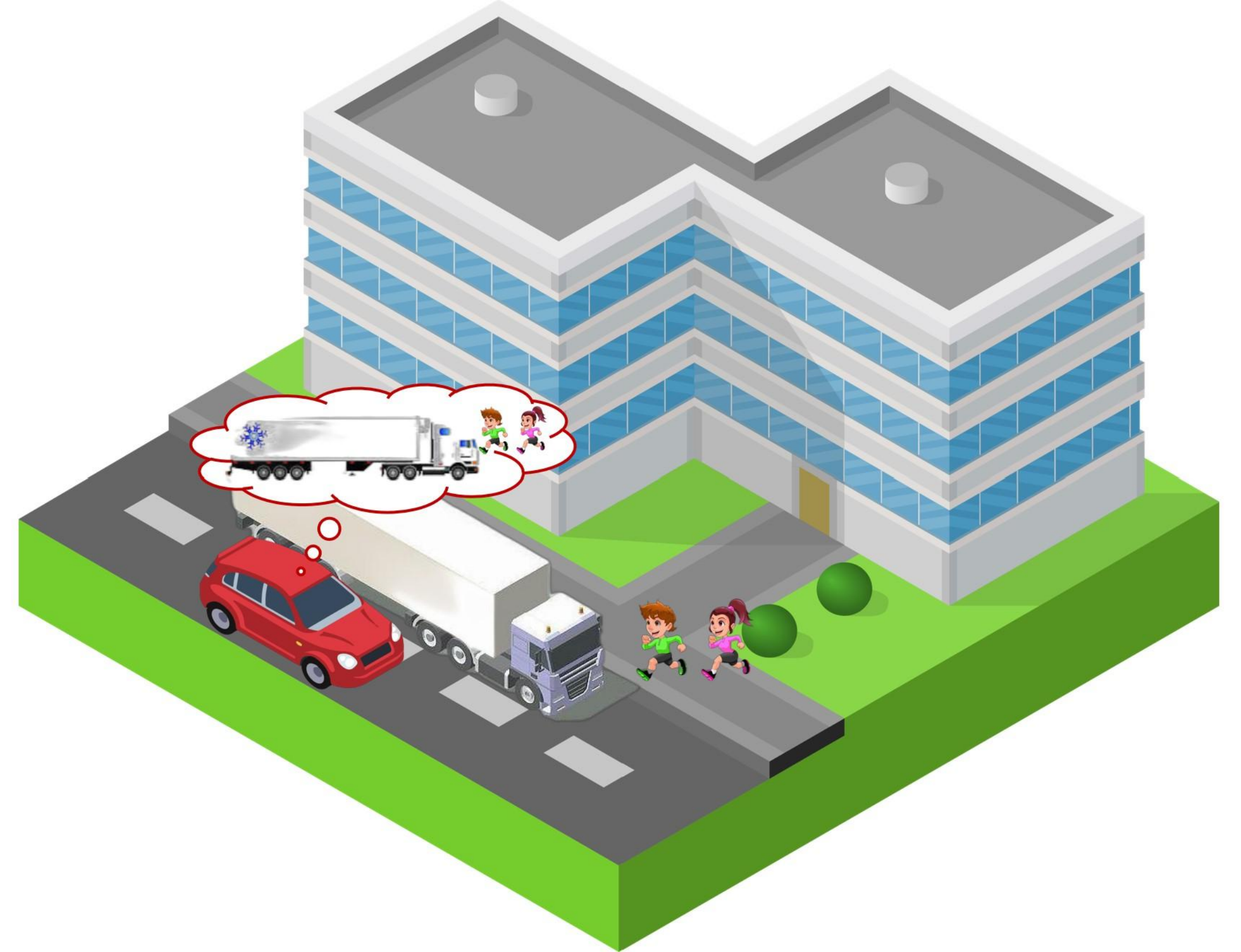}}
\caption{An illustration of the advantages of collaborative perception over stand-alone intelligence}
\label{fig:intro}
\end{figure}

One major bottleneck of achieving ultra-reliability in autonomous driving is the fundamental defect of stand-alone intelligence due to the single perception viewpoint. As illustrated in Fig.~\ref{fig:intro}(a), the autonomous vehicle could not detect the pedestrians in its blind zone caused by the truck, which may lead to a severe accident. Great efforts have been put into single-vehicle multi-view object detection with multiple heterogeneous sensors~\cite{chen2017multi,deng2019mlod} or homogeneous sensors~\cite{li2019stereo,wang2022detr3d}, but the intrinsic limitation of stand-alone intelligence still exists.

\vspace{-0.3cm}

Thanks to the Vehicle-to-Everything (V2X) network~\cite{zeadally2020vehicular}, {\it interconnected autonomous driving} is a highly-anticipated solution to occlusions, and thus enables advanced autonomous driving capabilities in complex scenarios such as intersections and overtaking. A vehicle can exchange the local sensor data with other terminals, including other vehicles and Road Side Units (RSUs), and then performs the object detection by fusing data from multiple viewpoints. The shared sensor data might contain information about the object in the blind zones of the ego vehicle, potentially enhancing the perception reliability~\cite{yang2021machine} (as shown in Fig.~\ref{fig:intro}(b)(c)). This procedure is named as {\it collaborative perception}, which can be categorized into three levels: raw-level (early fusion, e.g.~\cite{chen2019cooper}), feature-level (middle fusion, e.g.~\cite{fcooper,wang2020v2vnet}), and object-level (late fusion, e.g.~\cite{kim2015impact}).

\begin{figure}
\centering
\includegraphics[height=7cm]{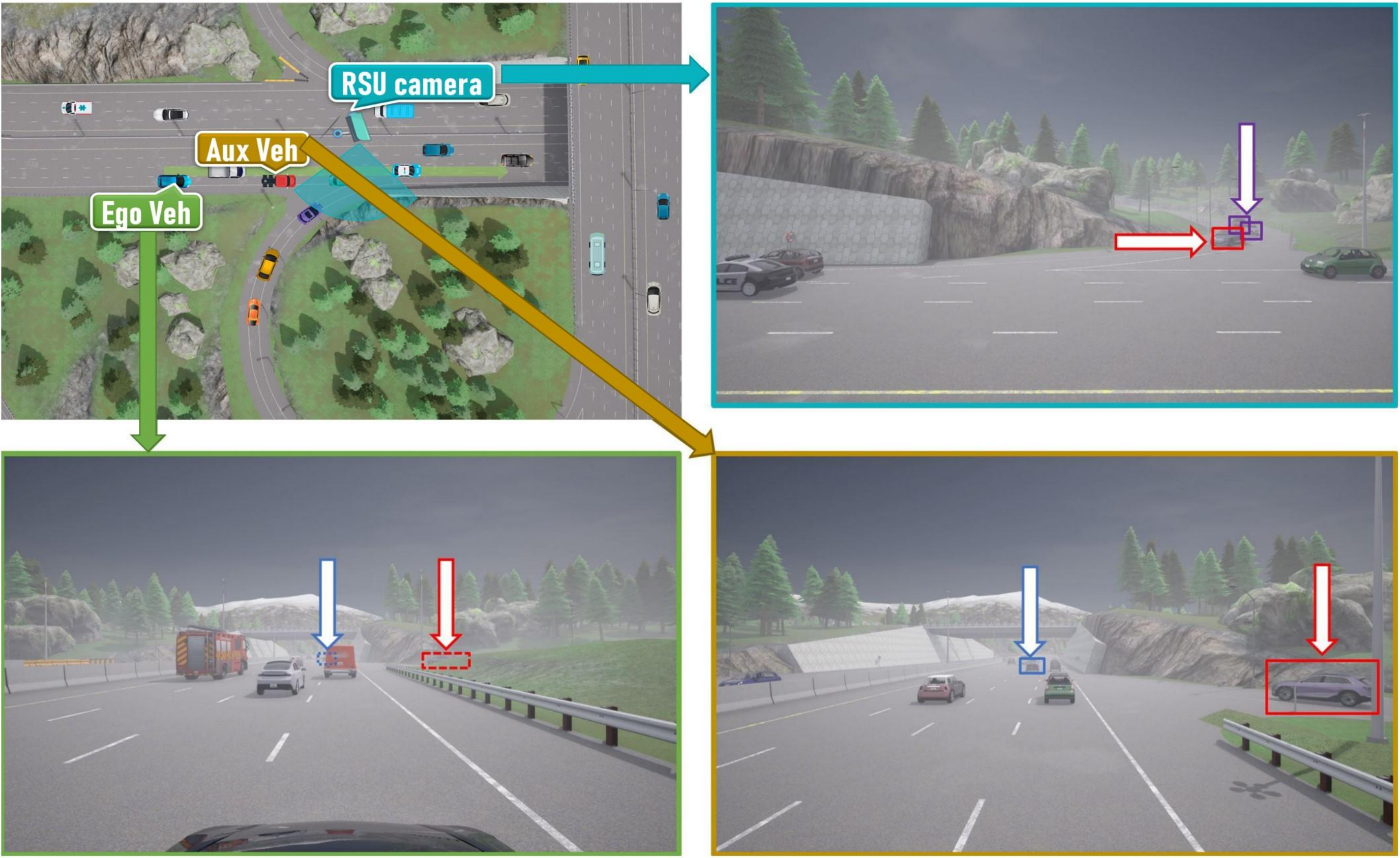}
\caption{An example of multi-view object detection in DOLPHINS dataset. There is a right merging lane in front of the ego vehicle. Because of the occlusion, the ego vehicle can hardly detect the purple vehicle ({\it red box}) on the branch and the police car ({\it blue box}). The auxiliary vehicle is in front of the ego vehicle, which can see both object vehicles distinctly. Additionally, the RSU can detect another two vehicles ({\it purple box}) on the branch}
\label{fig:sample}
\end{figure}

However, the lack of large-scale datasets for collaborative autonomous driving has been seriously restricting the research of collaborative perception algorithms. Traditional datasets focus on a single viewpoint, i.e., the ego vehicle. In the past decade, KITTI~\cite{Geiger2012CVPR}, nuScenes~\cite{caesar2020nuscenes}, and Waymo Open~\cite{sun2020scalability} have successfully accelerated the development of stand-alone self-driving algorithms with a huge amount of multi-modality data. But all of the information is collected from the ego vehicle view. Unfortunately, the most challenging but the greatest beneficial issue is the large parallax due to strong perspective changes between different terminals, i.e., aux vehicles and RSUs, as illustrated in Fig.~\ref{fig:sample}. The large parallax leads to various occlusion relationships between objects, which may help the terminals to fulfill the blind zones, but also put forward the matching of the same object from different perspectives. Recently, some pioneer works have concentrated on datasets with multiple viewpoints, such as OPV2V~\cite{opv2v}, V2X-Sim~\cite{v2x-sim}, and DAIR-V2X~\cite{yu2022dair}. Nevertheless, either data from aux vehicles (Vehicle-to-Vehicle, V2V) and RSUs (Vehicle-to-Infrastructure, V2I) are not provided simultaneously, or only an intersection scenario is considered. A more comprehensive dataset is required to fully support the development of V2X-based collaborative autonomous driving algorithms.

\begin{figure}
\centering
\resizebox{0.8\linewidth}{!}{
\begin{tikzpicture}[font=\tiny]
  \tkzRadarDiagramFromFile[
        scale=.4,
        label distance=.5cm,
        gap     = 1,
        label space= 0.8,  
        lattice = 5]{radar7.dat}
  \tkzRadarLegendFromFile[thick,                        
                          mark       = ball,
                          mark size  = 2pt,
                          fill       = green!40]{radar7.dat}   
\end{tikzpicture}
}
\caption{A comparison with 3 brand new collaborative perception datasets: OPV2V~\cite{opv2v}, V2X-Sim~\cite{v2x-sim}, and DAIR-V2X-C~\cite{yu2022dair}, as well as 2 well-known single-vehicle autonomous driving datasets: KITTI~\cite{Geiger2012CVPR} and nuScenes~\cite{caesar2020nuscenes}. A detailed comparison is provided in Sec.~\ref{sec:related}}
\label{fig:radar}
\end{figure}
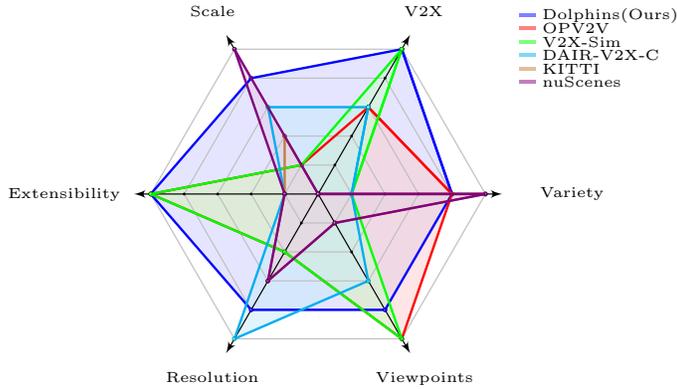

To meet the demands, we present {\bf DOLPHINS}, a new {\bf D}ataset for c{\bf OL}labor-ative {\bf P}erception enabled {\bf H}armonious and {\bf I}nterconnected {\bf S}elf-driving. We use the CARLA simulator~\cite{Dosovitskiy17} to complete this work, which can provide us with realistic environment modeling and real-time simulations of the dynamics and sensors of various vehicles. Fig.~\ref{fig:radar} briefly demonstrates the advantages of DOLPHINS in six dimensions.

\begin{figure}
\centering
\subfigure[Ego vehicle]{
\begin{minipage}{0.31\linewidth}
\centering
\includegraphics[width=\linewidth]{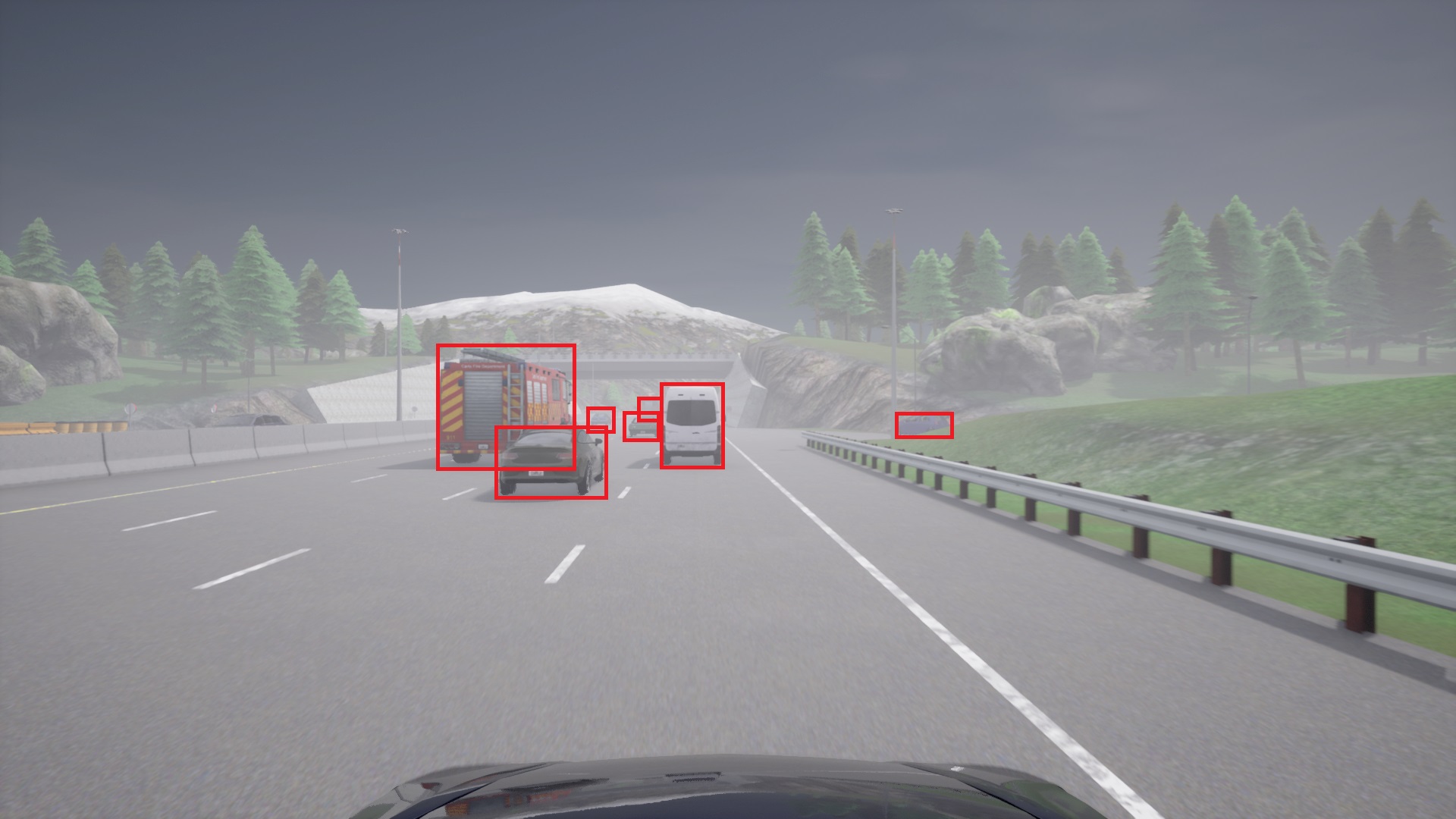}\\
\vspace{0.02cm}
\includegraphics[width=\linewidth]{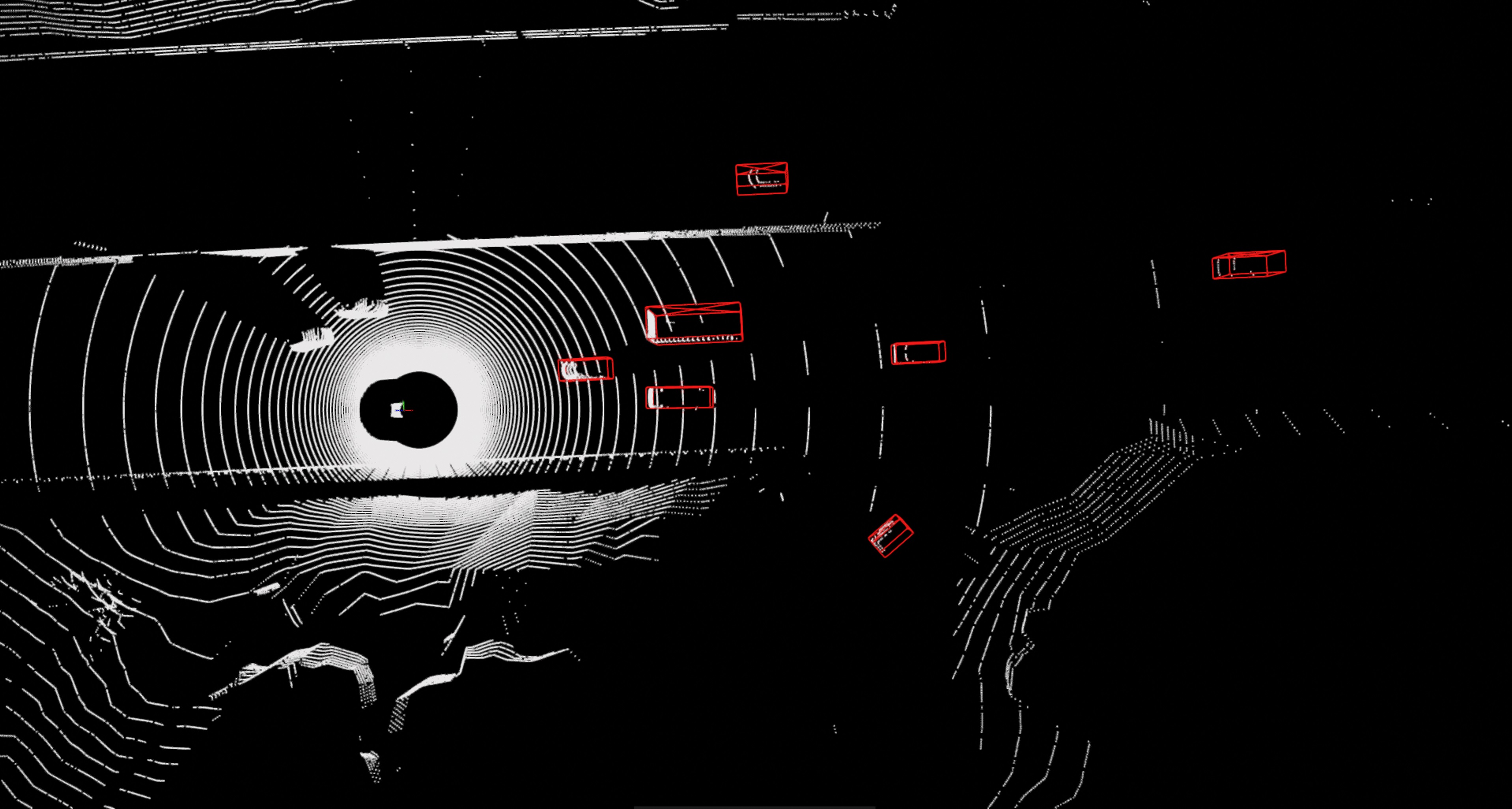}\\
\vspace{0.02cm}
\end{minipage}
}
\subfigure[RSU]{
\begin{minipage}{0.31\linewidth}
\centering
\includegraphics[width=\linewidth]{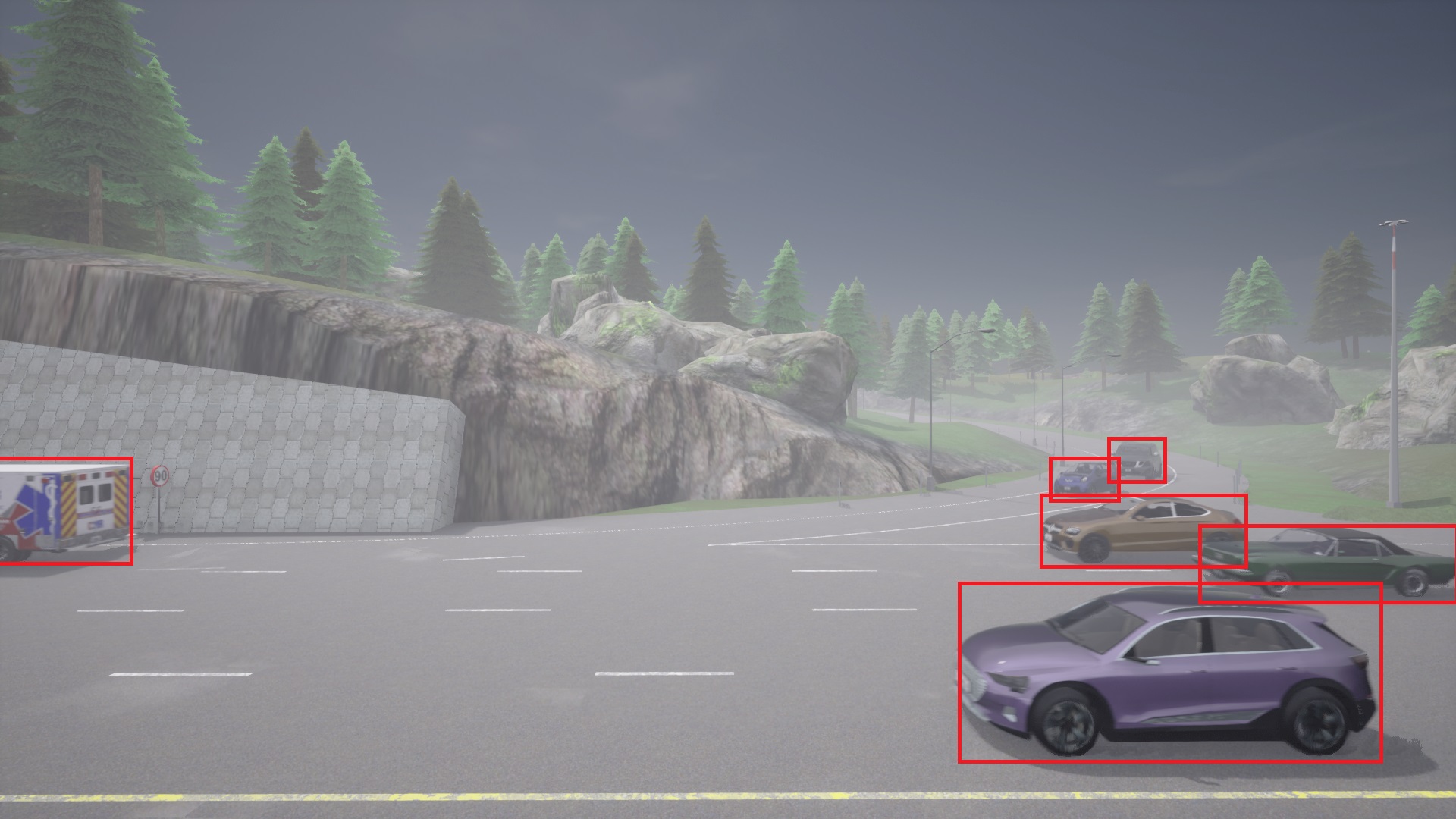}\\
\vspace{0.02cm}
\includegraphics[width=\linewidth]{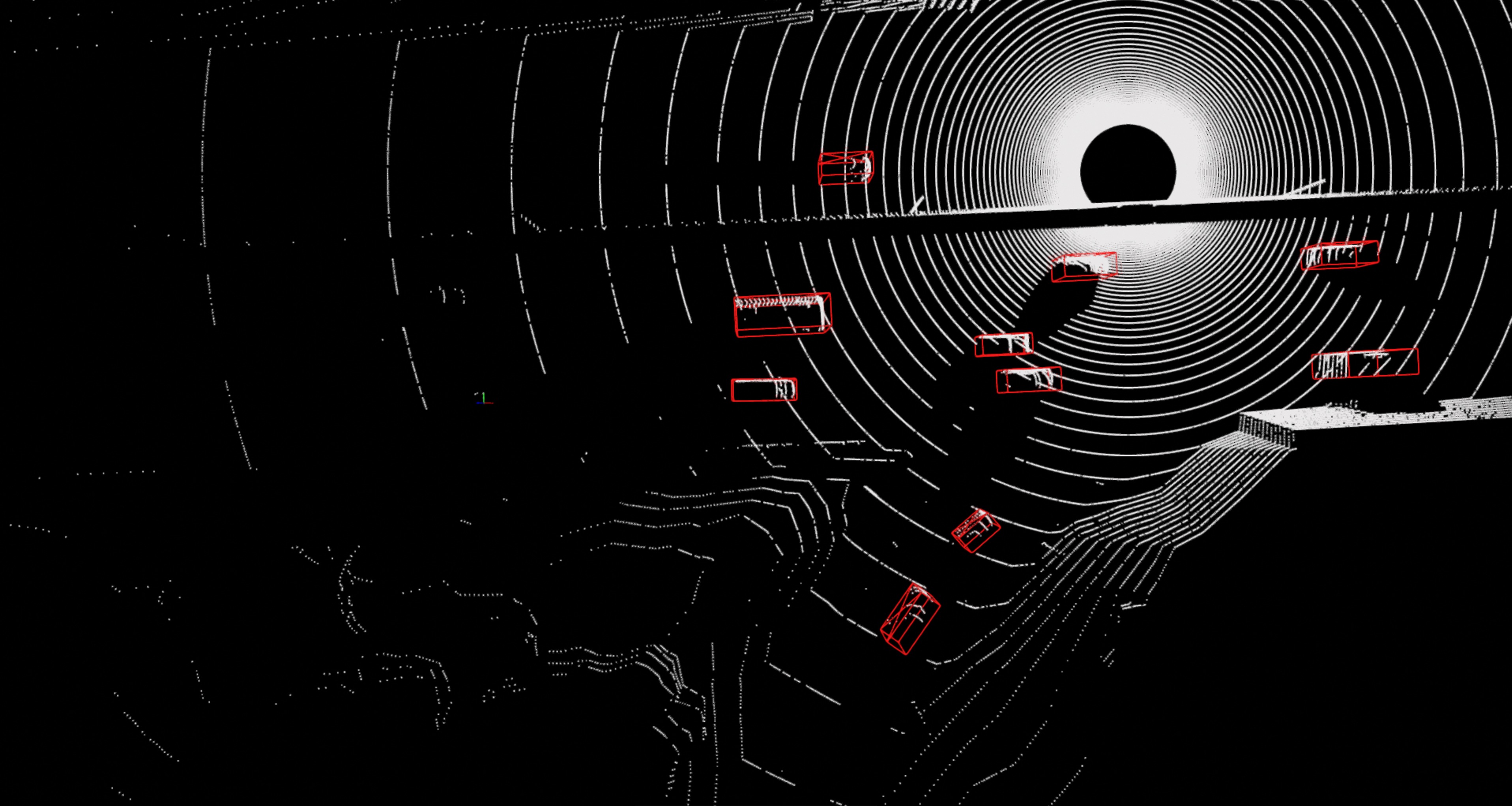}\\
\vspace{0.02cm}
\end{minipage}
}
\subfigure[Aux vehicle]{
\begin{minipage}{0.31\linewidth}
\centering
\includegraphics[width=\linewidth]{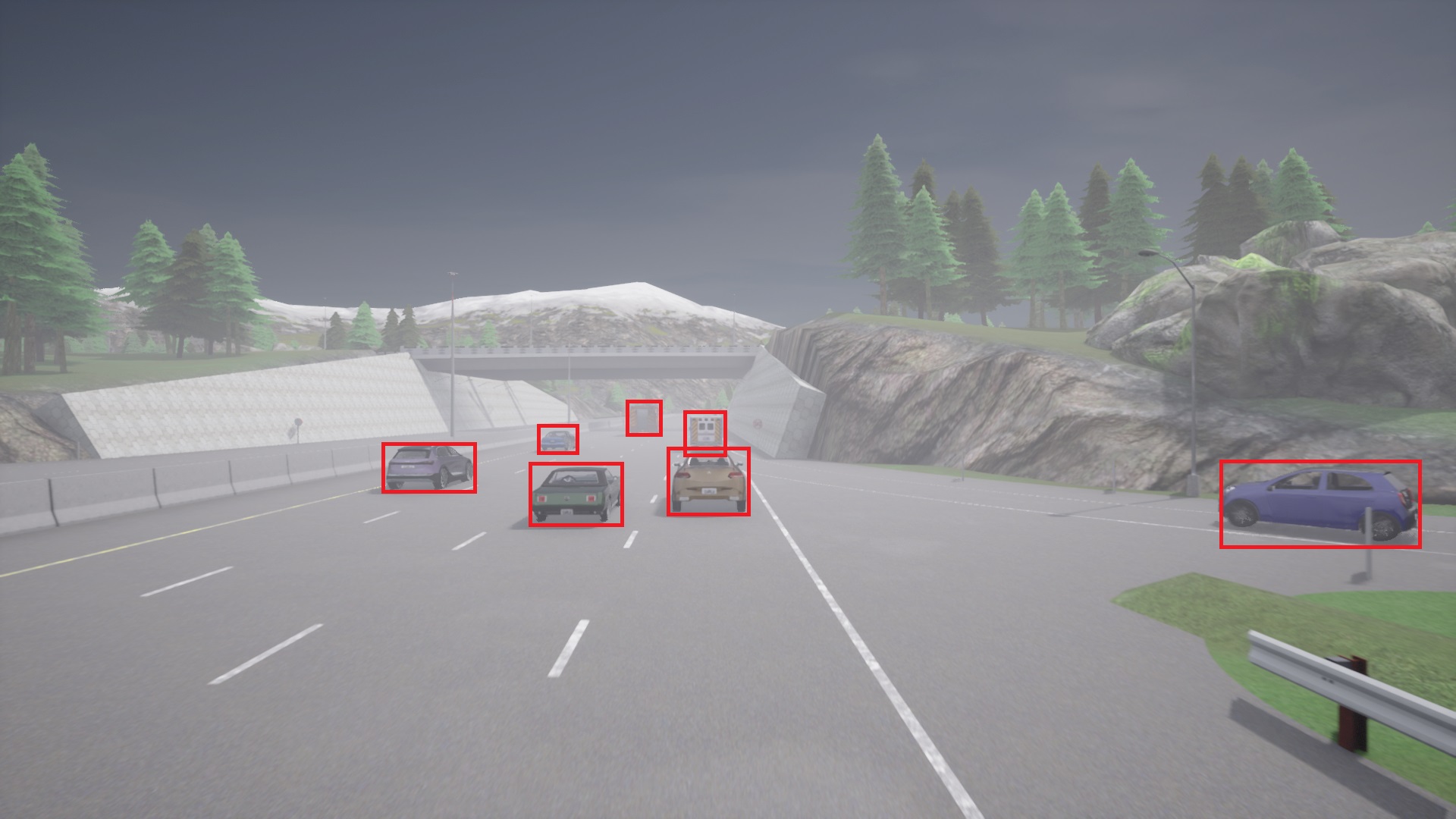}\\
\vspace{0.02cm}
\includegraphics[width=\linewidth]{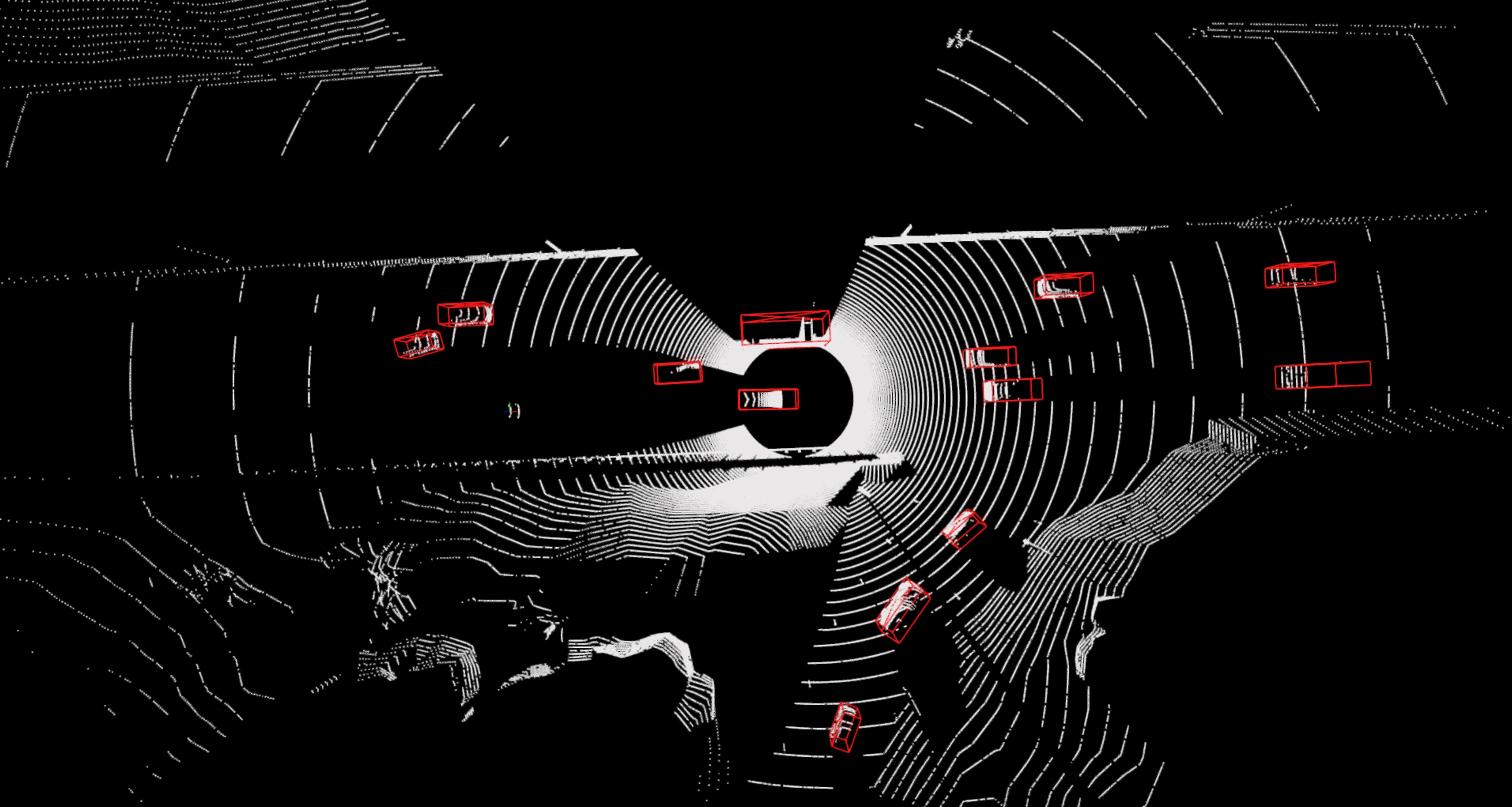}\\
\vspace{0.02cm}
\end{minipage}
}
\centering
\caption{An illustration of temporary-aligned images and point clouds from three viewpoints. The position of each viewpoint is demonstrated in Fig.~\ref{fig:sample}}
\label{fig:img&lidar}
\end{figure}

\begin{description}
\item[V2X] DOLPHINS contains \uline{temporally-aligned images and point clouds from both aux vehicles and RSUs simultaneously}, which provides a universal out-of-the-box benchmark platform for the development and verification of V2V and V2I enabled collaborative perception without extra generation of data.
\item[Variety] DOLPHINS includes \uline{6 typical autonomous driving scenarios}, which is second only to real-world single-vehicle datasets~\cite{Geiger2012CVPR,caesar2020nuscenes}. Our dataset includes urban intersections, T-junctions, steep ramps, highways on-ramps, and mountain roads, as well as dynamic weather conditions. Different scenarios raise different challenges to autonomous driving, such as dense traffic, ramp occlusions, and lane merging. More detailed information on traffic scenarios is presented in Sec.~\ref{sec:scenario}.
\item[Viewpoints] Considering the actual driving situation, \uline{3 different viewpoints are meticulously set for each scenario, including both RSUs and vehicles}. The data collected from viewpoints can achieve full coverage of key areas in each scenario as illustrated in Fig.~\ref{fig:img&lidar}, which ensures each object appears in at least one perspective and is thus sufficient for the validation of collaborative perception algorithms. More specific locations of each viewpoint are illustrated in Fig.~\ref{fig:scenario}.
\item[Scale] In total, temporally-aligned images and point clouds are recorded over \uline{42376 frames from each viewpoint}, which is much larger than any other dataset for collaborative perception. \uline{3D information of 292549 objects is annotated in KITTI format} for ease of use, along with the geo-positions and calibrations. Statistical analysis of objects is provided in Sec.~\ref{sec:analysis}.
\item[Resolution] DOLPHINS furnishes high-resolution images and point clouds to maintain sufficient details. \uline{Full-HD ($1920\times1080$) cameras and 64-line LiDARs} equipped on both vehicles and RSUs, which are both among the highest quality in all datasets. Detailed descriptions of sensors are stated in Sec.~\ref{sec:sensor}.
\item[Extensibility] We also release the related codes of DOLPHINS, which contains the well-organized API to help researchers to generate additional data on demand, which makes DOLPHINS easily extensible and highly flexible.
\end{description}

We also conduct a comprehensive benchmark of state-of-the-art algorithms on DOLPHINS. Three typical tasks are considered: 2D object detection, 3D object detection, and multi-view collaborative perception. Other tasks, such as tracking, are also supported in DOLPHINS but not exhibited here. Besides, we construct two raw-level fusion schemes: the point clouds from the ego vehicle and the other viewpoint, and the image from the ego vehicle and point clouds from the RSU. The results of the raw-level fusion algorithms reveal the dual character of interconnected self-driving: enhancing the precision with more information or reducing the cost of sensors on the self-driving vehicles within the same precision.

As a new {\it large-scale various-scenario multi-view multi-modality} dataset, we hope this work brings a new platform to discover the potential benefits of connected intelligence. Our main contributions are summarized as:

\begin{enumerate}[i.]
    \item release DOLPHINS dataset with different scenarios, multiple viewpoints, and multi-modal sensors, aiming to inspire the research of collaborative autonomous driving;
    \item provide open source codes for on-demand generation of data;
    \item benchmark several state-of-the-art methods in 2D object detection, 3D object detection, and multi-view collaborative perception, illustrating the possibility of solving blind zones caused by occlusions as well as cutting the cost of self-driving vehicles by V2V and V2I communication.
    
\end{enumerate}

\section{Related works}
\label{sec:related}

There are many relative research areas, such as object detection, collaborative perception, and autonomous driving dataset. Due to the space limitation, some representative works which inspire us are introduced here, and the differences with our proposed dataset are highlighted.\newline

\noindent{\bf Object detection} is one of the most important tasks in autonomous driving. Typically, there are two kinds of object detectors, distinguished by whether to generate region proposals before the object detection and bounding box regression. R-CNN family~\cite{girshick2014rich,Girshick_2015_ICCV,ren2015faster,he2017mask} is the representative of two-stage detectors, which exhibits epoch-making performance. On the other hand, the single-stage detectors, such as SSD~\cite{liu2016ssd} and YOLO~\cite{redmon2016you,redmon2017yolo9000,redmon2018yolov3}, focus on the inference time and perform significantly faster than the two-stage competitors. Recently, CenterNet~\cite{duan2019centernet} and CornerNet~\cite{law2018cornernet} propose a new detection method without anchor generation. They directly predict the key points per-pixel, which makes the detection pipeline much simpler. DETR~\cite{detr} firstly brings transformer architecture into object detection tasks.\newline

\noindent{\bf Collaborative perception} is a growing topic in the intelligent transportation society. Due to the 3D information provided by point clouds, LiDAR-based data fusion and object detection have been widely discussed. \cite{chen2019cooper} proposes a raw-level fusion on point clouds with a deep network for detection. \cite{rawashdeh2018collaborative,xiao2018multimedia} aim to use deep neural networks to enhance the perception outputs for sharing. V2VNet~\cite{wang2020v2vnet} considers the feature-level fusion and uses a compressor to reduce the size of original point clouds, which is vital in bandwidth-limited V2X communications. V2V-ViT~\cite{v2x-vit} introduces vision transformer to conquer the noises from communication. \cite{nassar2019simultaneous} and \cite{nassar2020geograph} consider the pure image fusion on feature-level and conduct a re-identification task during the detection. The latter uses Graph Neural Network (GNN) for perception and clustering.\newline

\noindent{\bf Autonomous driving datasets} are the key to evaluating the performance of detection methods. The commonly used KITTI~\cite{Geiger2012CVPR} and nuScenes~\cite{caesar2020nuscenes} only contain data from the ego vehicle. Pasadena Multi-view ReID dataset is proposed in~\cite{nassar2019simultaneous}, which contains data from different viewpoints of a single object. However, the objects are only street trees, which is not enough for autonomous driving. OPV2V~\cite{opv2v}  uses CARLA simulator~\cite{Dosovitskiy17} to produce multi-view autonomous driving data, but it only considers V2V communication. V2X-Sim~\cite{v2x-sim} is also a CARLA-based simulated dataset. The first version of V2X-Sim only contains point clouds from different vehicles, which can only be applied for V2V communication. The second version of V2X-Sim contains both RGB images and the infrastructure viewpoints. Nevertheless, it still only considers the intersections scenario, and the BEV Lidar on the infrastructure is not realistic. By late February 2022, a new real-world connected autonomous driving dataset DAIR-V2X~\cite{yu2022dair} is released. It consists of images and point clouds from one vehicle and one RSU, and contains both high-ways and intersections. However, DAIR-V2X is not capable of V2V data fusion or any other scenarios with more than two terminals. Our proposed dataset is generated by the CARLA simulator with six different scenarios and reasonable settings of RSUs and aux vehicles. Besides, with the related codes (which will also be released with the dataset), researchers can add any type and any number of sensors at any location as needed. A comparison to the above datasets is provided in Table~\ref{tab:dataset-compare}.

\section{DOLPHINS Dataset}
\label{dataset}

\subsection{Settings of traffic scenarios}
\label{sec:scenario}

\begin{figure}
\centering
\subfigure[Scenario 1 (7046 frames)]{\includegraphics[width=4cm]{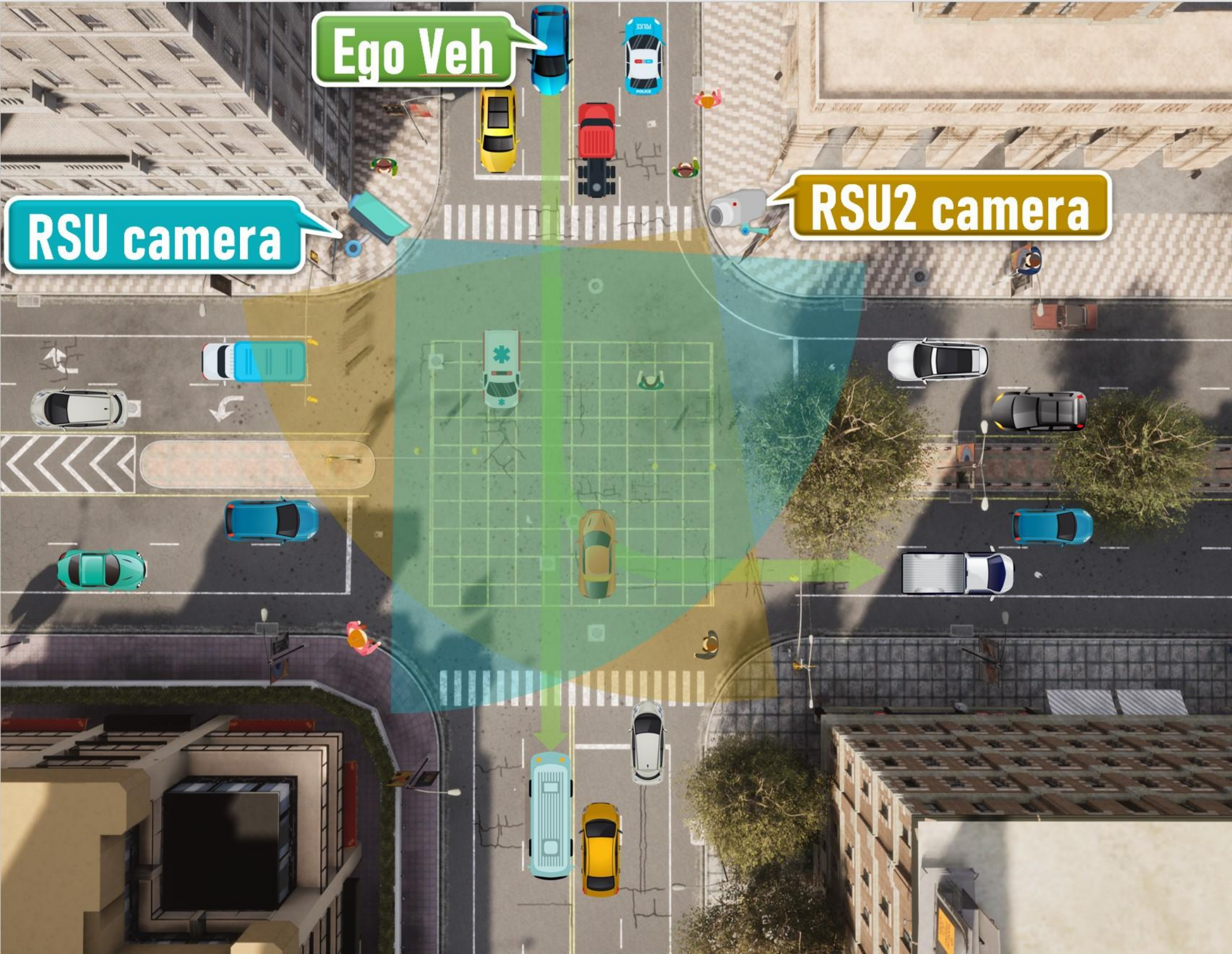}} 
\subfigure[Scenario 2 (7020 frames)]{\includegraphics[width=4cm]{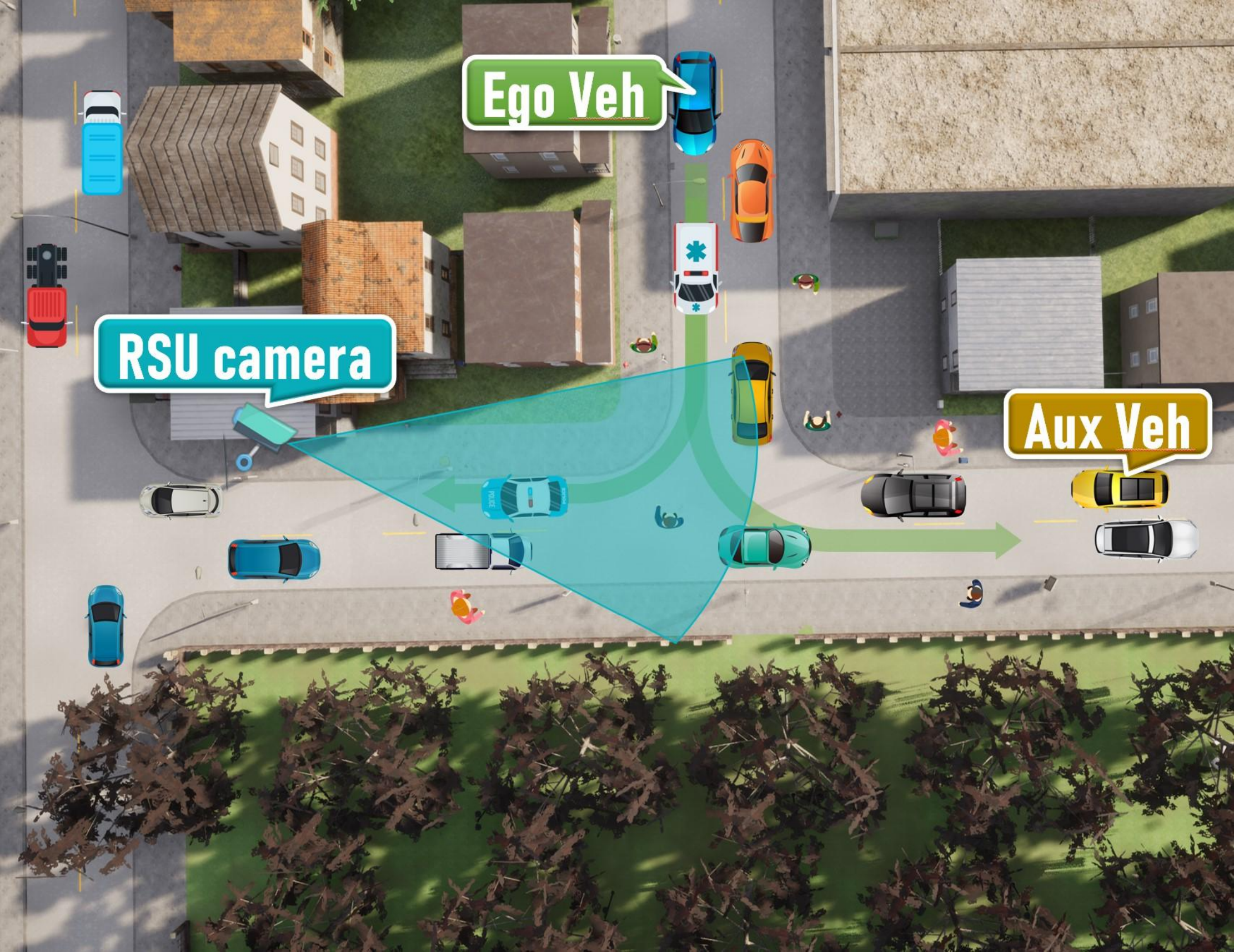}}
\subfigure[Scenario 3 (7043 frames)]{\includegraphics[width=4cm]{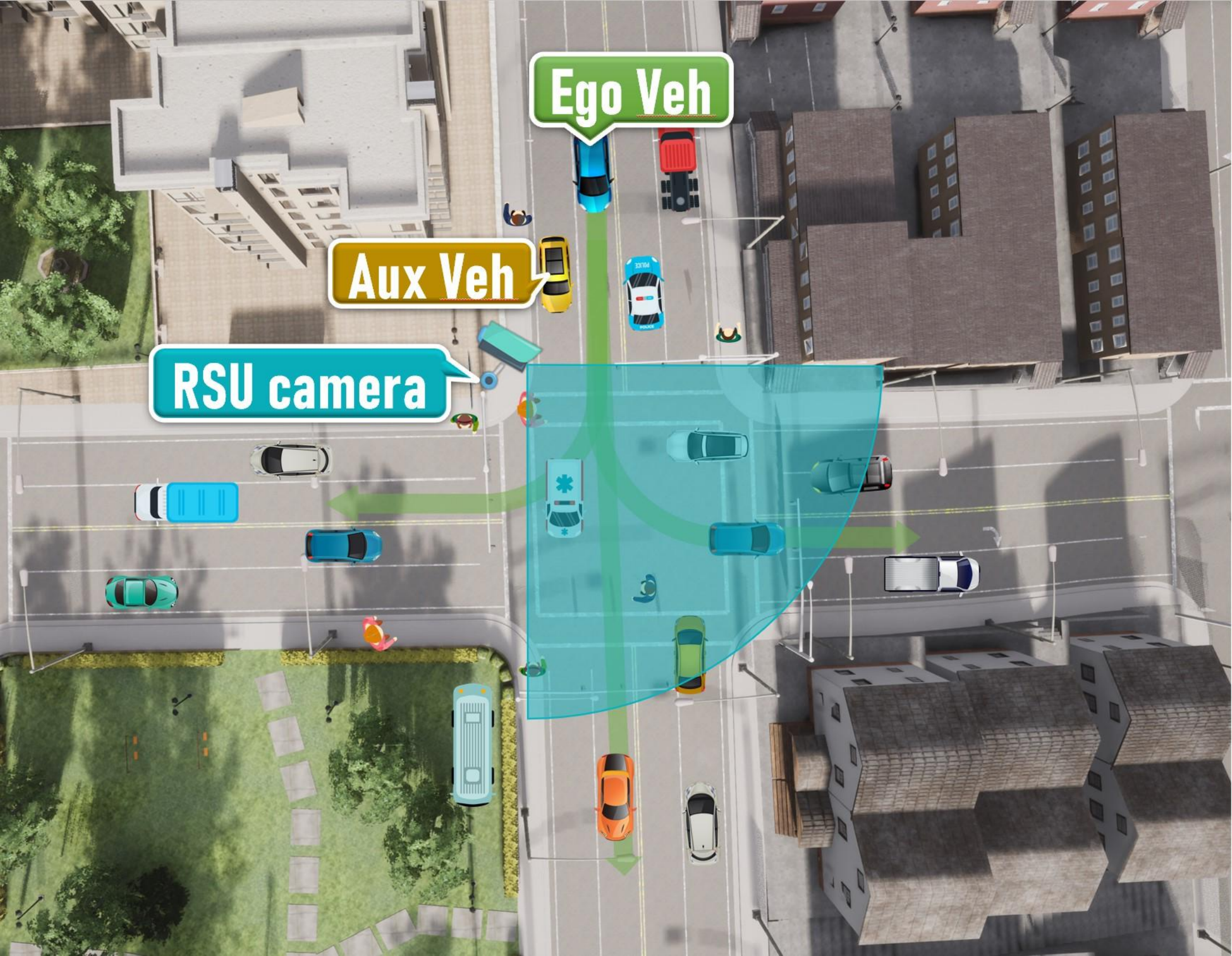}}
\\
\centering
\subfigure[Scenario 4 (7057 frames)]{\includegraphics[width=4cm]{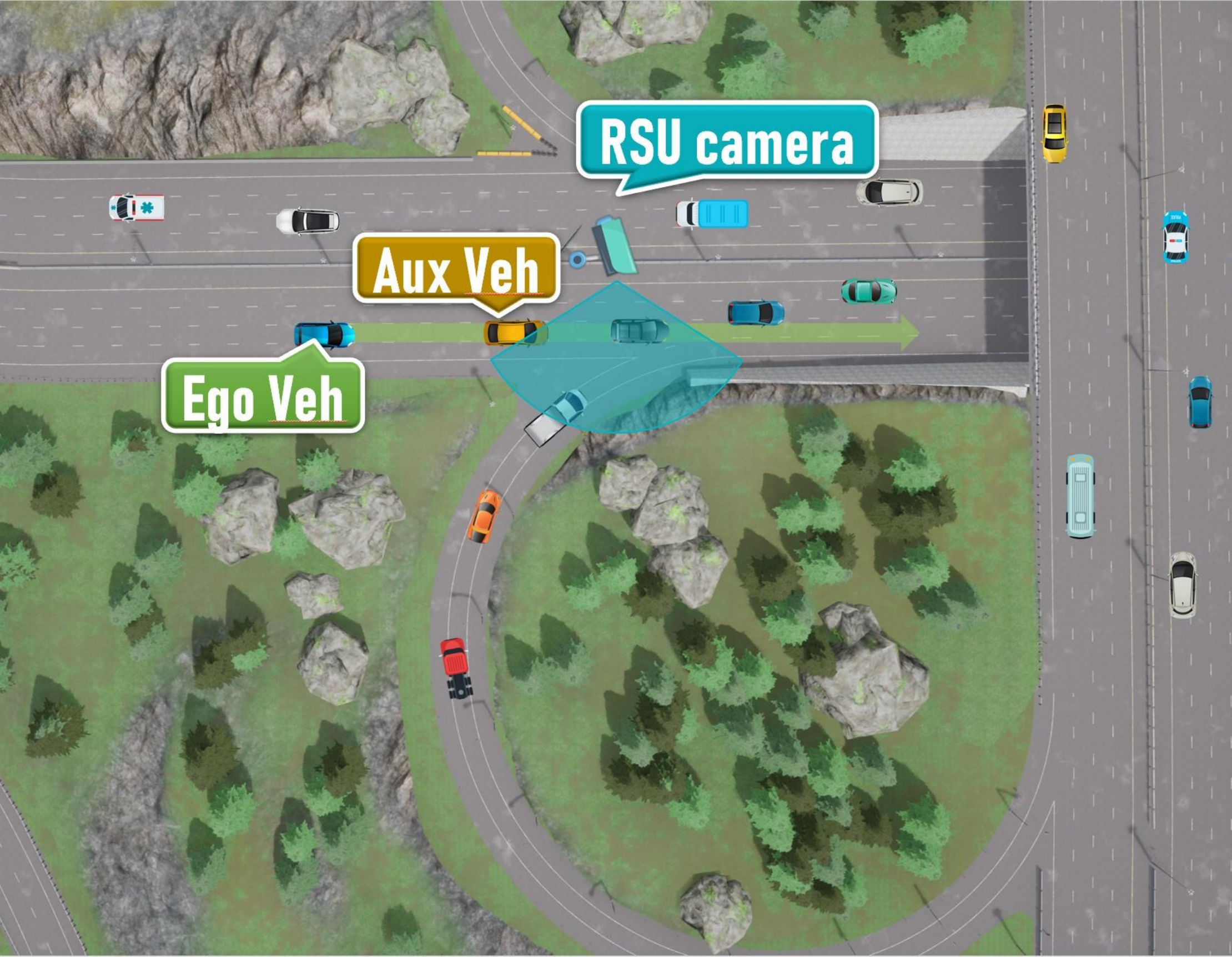}}
\subfigure[Scenario 5 (7011 frames)]{\includegraphics[width=4cm]{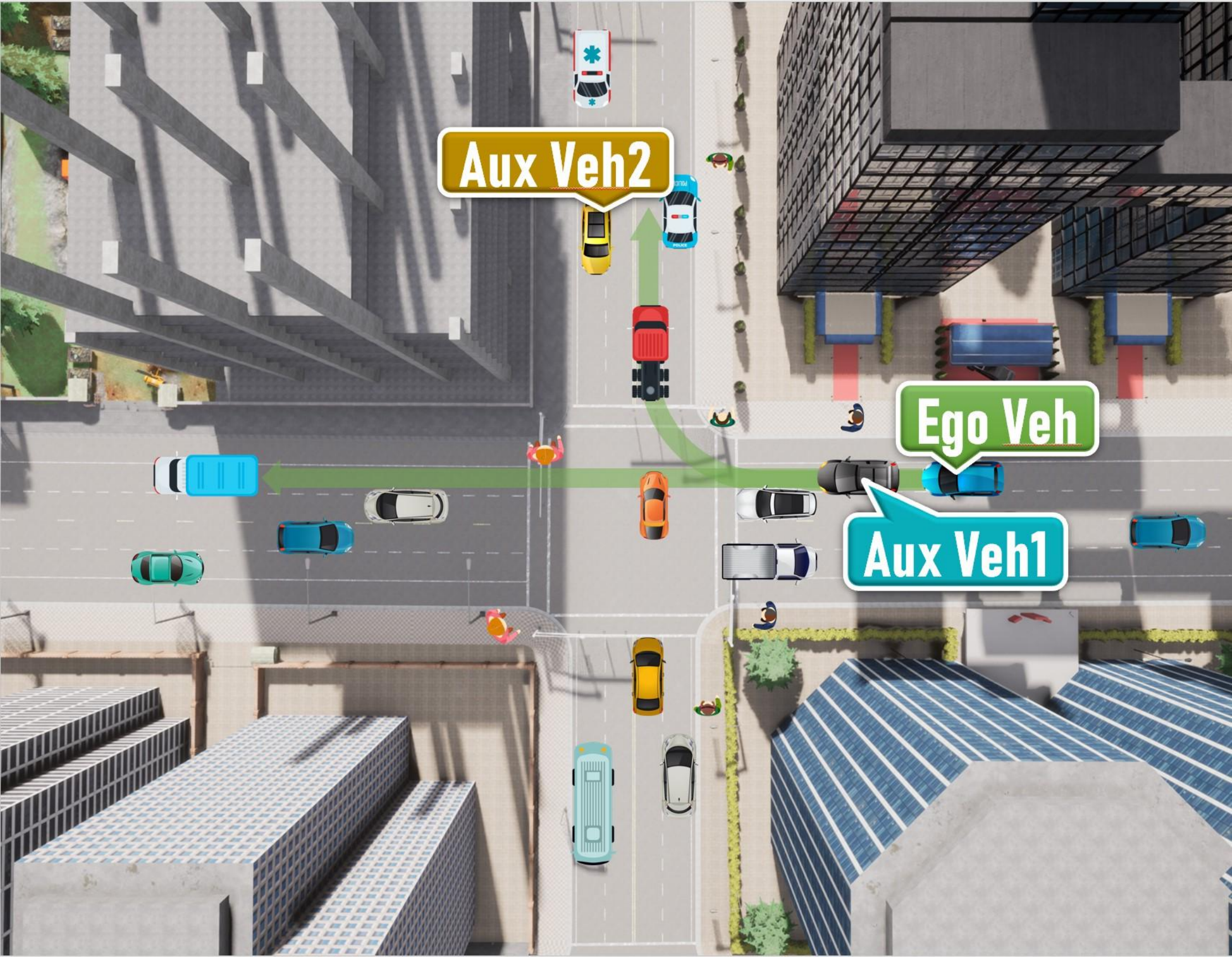}}
\subfigure[Scenario 6 (7199 frames)]{\includegraphics[width=4cm]{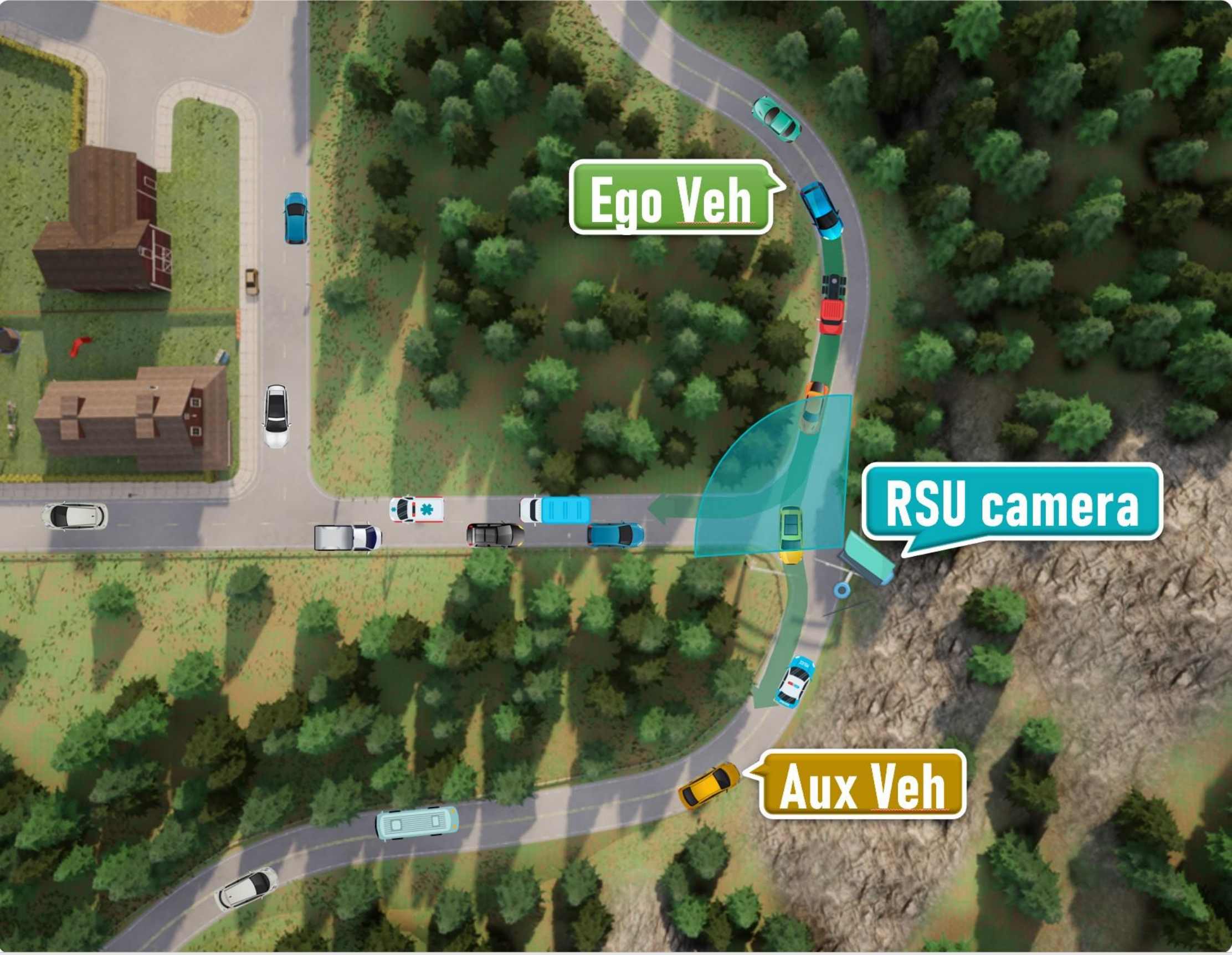}}
\caption{All ego vehicles are driving along a pre-defined route ({\it green arrows}), while each RSU camera is settled with a fixed direction and range ({\it blue or brown sector mark}). We also mark positions where the ego vehicle or possible auxiliary vehicles are initialized. Among all scenarios, (a) and (e) are two intersection scenarios; (b) is the scenario of a T-junction with moderate rain; (c) is also a crossroads while the ego vehicle is on a steep ramp; (d) is a scenario existing a right merging lane on the expressway, and the weather is foggy; (f) is the scenario of a mountain road. All scenarios have plenty of occlusion situations}
\label{fig:scenario}
\end{figure}

We select six typical autonomous driving scenarios and several common types of weather from the preset scenarios of the CARLA simulator (as shown in Fig.~\ref{fig:scenario}). In each scenario, we set three units (RSU or vehicles) to collect both images and point clouds information. The first unit is attached to the vehicle we drive, namely, the ego vehicle, which provides us with the main viewpoint. In each simulation round, we initialized it at a specific location. The other two units will also be set up at appropriate positions. They are set on the RSUs or the auxiliary vehicles selected from the scenario and initialized at a specially designated point with a stochastic vehicle model.

\begin{table}
\centering
\caption{A detailed comparison between datasets. For DAIR-V2X, we choose DAIR-V2X-C since only this part is captured synchronously by both vehicles and infrastructure sensors}
\label{tab:dataset-compare}
\resizebox{\textwidth}{!}{%
\begin{tabular}{@{}cccccccc@{}}
\toprule
Dataset                          & Year                  & V2X                      & Scenarios          & Viewpoints                    & Frames                  & Extensibility               & Resolution                \\ \midrule
\multirow{2}{*}{KITTI}           & \multirow{2}{*}{2012} & \multirow{2}{*}{none}    & \multirow{2}{*}{-} & \multirow{2}{*}{1}            & \multirow{2}{*}{15 k}   & \multirow{2}{*}{×}          & 1382$\times$512           \\
                                 &                       &                          &                    &                               &                         &                             & 64 lines                  \\
\multirow{2}{*}{nuScenes}        & \multirow{2}{*}{2019} & \multirow{2}{*}{none}    & \multirow{2}{*}{-} & \multirow{2}{*}{1}            & \multirow{2}{*}{1.4 M}  & \multirow{2}{*}{×}          & 1600$\times$1200          \\
                                 &                       &                          &                    &                               &                         &                             & 32 lines                  \\
\multirow{2}{*}{OPV2V}           & \multirow{2}{*}{2021} & \multirow{2}{*}{V2V}     & \multirow{2}{*}{6} & \multirow{2}{*}{2-7 (avg. 3)} & \multirow{2}{*}{11.5 k} & \multirow{2}{*}{\checkmark} & 800$\times$600            \\
                                 &                       &                          &                    &                               &                         &                             & 64 lines                  \\
\multirow{2}{*}{V2X-Sim}         & \multirow{2}{*}{2022} & \multirow{2}{*}{V2V+V2I} & \multirow{2}{*}{1} & \multirow{2}{*}{2-5}          & \multirow{2}{*}{10 k}   & \multirow{2}{*}{\checkmark} & 1600$\times$900           \\
                                 &                       &                          &                    &                               &                         &                             & 32 lines                  \\
\multirow{2}{*}{DAIR-V2X-C}      & \multirow{2}{*}{2022} & \multirow{2}{*}{V2I}     & \multirow{2}{*}{1} & \multirow{2}{*}{2}            & \multirow{2}{*}{39 k}   & \multirow{2}{*}{×}          & 1920$\times$1080          \\
                                 &                       &                          &                    &                               &                         &                             & I: 300 lines; V: 40 lines \\
\multirow{2}{*}{DOLPHINS (Ours)} & \multirow{2}{*}{2022} & \multirow{2}{*}{V2V+V2I} & \multirow{2}{*}{6} & \multirow{2}{*}{3}            & \multirow{2}{*}{42 k}   & \multirow{2}{*}{\checkmark} & 1920$\times$1080          \\
                                 &                       &                          &                    &                               &                         &                             & 64 lines                  \\ \bottomrule 
\end{tabular}%
}
\end{table}
\vspace{-3mm}

In each scenario, our ego vehicle chooses a specific route. At the same time, we collect the information of all sensors synchronously every 0.5 seconds in the simulation environment, i.e., at the rate of 2 fps. After the vehicle passes through the specific scenario, we wind up the current simulation round, reinitialize the scenario and start a new one. During each round, except for our ego vehicle and the possible auxiliary vehicle, all other traffic participants appear in a reasonable position randomly at the beginning and choose their route by themselves freely.

\subsection{Settings of sensors}
\label{sec:sensor}

\begin{figure}
\centering
\includegraphics[width=\linewidth]{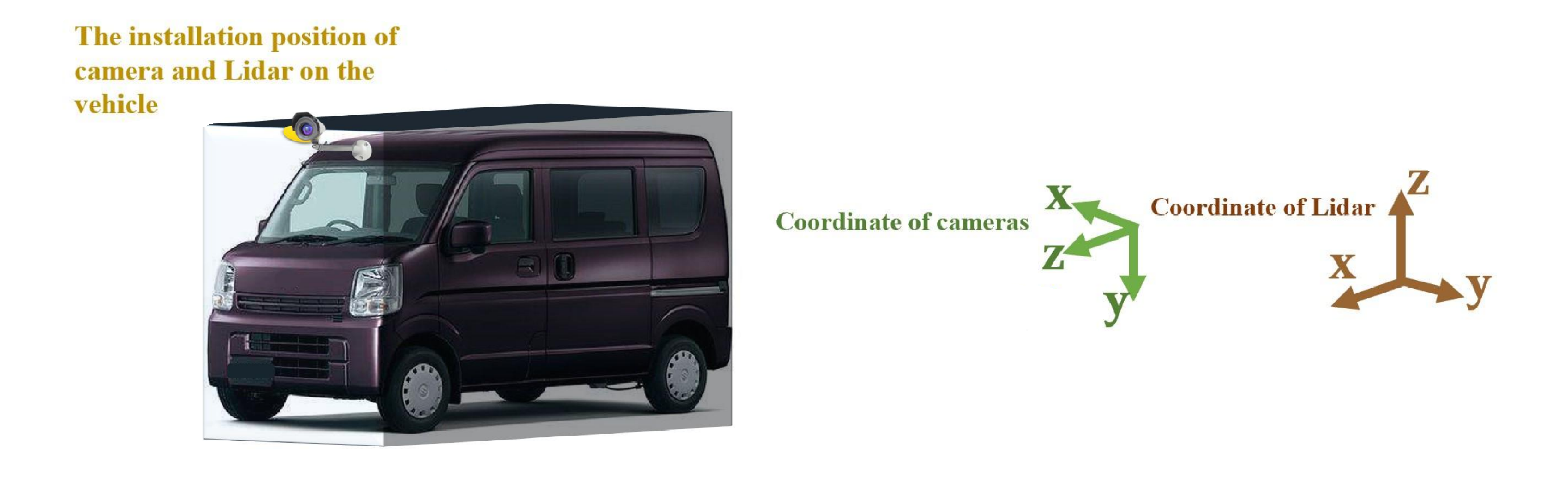}
\caption{Sketch map of a fully equipped unit (take a vehicle as an example) and the coordinate system of each sensor}
\label{fig:sensor}
\end{figure}

We equip each unit with a LiDAR and an RGB camera, whose parameters are listed in Table~\ref{table:parameter}. For the convenience of calibration between different sensors, we install both camera and LiDAR on the same point. The position of sensors on the vehicle is illustrated in Fig.~\ref{fig:sensor}.

\vspace{-0.5cm}

\subsection{Extra data and calibrations}
\label{sec:extra_calib}

For each scenario, We divide our data into the training set and the test set at the ratio of 8:2. Each set contains the original pictures taken by the camera, the point cloud information generated by LiDAR, and the ground truth labels, and the calibration files. The labels include the following information: (\rmnum{1}) 2D bounding box of the object in the image, (\rmnum{2}) 3D object dimensions and location, (\rmnum{3}) the value of \normalem{\emph{alpha}} and \normalem{\emph{rotation\_y}} which are defined in the KITTI Vision Benchmark~\cite{Geiger2012CVPR}. Except for the above data, we further introduce two extra pieces of information in DOLPHINS: the locations of key vehicles and the context-aware labels. These two kinds of data are essential for collaborative autonomous driving. The geo-positions of vehicles can greatly help to align the perceptual information from different perspectives through coordinate transformations. Actually, to the best of our knowledge, all the published multi-view collaborative perception algorithms are based on the locations of each vehicle, no matter image-based~\cite{nassar2020geograph,nassar2019simultaneous} or LiDAR-based~\cite{chen2019cooper,fcooper,wang2020v2vnet}. Besides, the interconnected autonomous vehicles can have wider perception fields with the help of other transportation participants and the RSUs, which means they can detect invisible objects. Most of the datasets only provide the labels of those who are in the view angle of sensors, which is not enough for the vehicles to make safe and timely decisions. We provide the labels of all traffic participants within 100 meters in front of or behind the ego unit, as well as 40 meters in the left and right side directions.
\vspace{-5mm}
\setlength{\tabcolsep}{4pt}
\begin{table}
\begin{center}
\caption{Parameters of sensors on different units}
\label{table:parameter}
\begin{tabular}{llll}
\bottomrule\noalign{\smallskip}
Sensor type & Parameter attributes & RSU & Vehicle\\
\noalign{\smallskip}
\bottomrule
\noalign{\smallskip}
\multirow{3}{*}{\makecell[l]{RGB\\Camera}} & Horizontal field of view in degrees & 90 & 90\\
 & Resolution & $1920\times1080$ & $1920\times1080$\\
 & Height in meters & 4 & 0.3+$h_{veh}$\footnotemark[1]\\
\noalign{\smallskip}
\hline
\noalign{\smallskip} 
\multirow{7}{*}{LiDAR} & Number of lasers & 64 & 64\\
 & Maximum distance to measure in meters & 200 & 200\\
 & Points generated by all lasers per second & $2.56\times10^{6}$ & $2.56\times10^{6}$\\
 & LiDAR rotation frequency & 20 & 20\\
 & Angle in degrees of the highest laser & 0 & 2\\
 & Angle in degrees of the lowest laser & -40 & -24.8\\
 & \makecell[l]{General proportion of points that\\ are randomly dropped} & 0.1 & 0.1\\
\bottomrule
\end{tabular}
\footnotetext[1] 01. $h_{veh}$ denotes the height of the ego vehicle
\end{center}
\end{table}
\setlength{\tabcolsep}{1.4pt}
\vspace{-1cm}

\subsection{Data analysis}
\label{sec:analysis}



To further analyze the data components of the dataset, we calculate the number of cars and pedestrians in each scenario both in the training dataset and the test dataset (as illustrated in Table~\ref{tab:object}). What's more, we categorize each object into three detection difficulty levels based on the number of laser points reflected by it in the point clouds. Easy objects reflect more than 16 points, as well as hard objects have no visible point, and the remaining objects are defined as moderate ones. In other words, the difficulty level actually indicates the occlusion level of each object. Since it is unlikely for us to manually annotate the occlusion level, such kind of definition is a suitable and convenient approximation. From the statistical analysis, it turns out that there is no pedestrian in scenarios 4 and 6, i.e., on high-way and mountain roads, which is self-consistent. Scenario 1 contains the most cars and pedestrians, as it is a crowded intersection. Scenario 2 is a T-junction, which has fewer directions for vehicles to travel. Scenario 3 is a steep ramp, which will be the hardest scenario along with Scenario 6, because of the severe occlusions caused by height difference.

\begin{table}
\centering
\caption{Statistical analysis of objects in DOLPHINS training and test dataset}
\label{tab:object}
\resizebox{\textwidth}{!}{%
\begin{tabular}{c|cccccc|cccccc}
\toprule
\multicolumn{1}{l|}{\multirow{3}{*}{Scenario}} & \multicolumn{6}{c|}{Training Dataset}                                             & \multicolumn{6}{c}{Test Dataset}                                               \\ \cline{2-13} 
\multicolumn{1}{l|}{}                          & \multicolumn{3}{c|}{Car}                       & \multicolumn{3}{c|}{Pedestrians} & \multicolumn{3}{c|}{Car}                     & \multicolumn{3}{c}{Pedestrians} \\ \cline{2-13} 
\multicolumn{1}{l|}{}                          & Easy   & Moderate & \multicolumn{1}{c|}{Hard}  & Easy     & Moderate    & Hard    & Easy  & Moderate & \multicolumn{1}{c|}{Hard} & Easy    & Moderate    & Hard    \\ \hline
1                                              & 27548  & 4423     & \multicolumn{1}{c|}{1090}  & 12370    & 2117        & 349     & 7048  & 1096     & \multicolumn{1}{c|}{296}  & 3079    & 579         & 96      \\
2                                              & 15428  & 1290     & \multicolumn{1}{c|}{567}   & 5641     & 2281        & 314     & 3895  & 330      & \multicolumn{1}{c|}{155}  & 1481    & 579         & 79      \\
3                                              & 14365  & 4029     & \multicolumn{1}{c|}{4291}  & 3003     & 3462        & 584     & 3631  & 1068     & \multicolumn{1}{c|}{1049} & 789     & 889         & 150     \\
4                                              & 34012  & 11771    & \multicolumn{1}{c|}{4089}  & 0        & 0           & 0       & 8497  & 2937     & \multicolumn{1}{c|}{1053} & 0       & 0           & 0       \\
5                                              & 31648  & 6201     & \multicolumn{1}{c|}{1993}  & 4797     & 9734        & 1476    & 7918  & 1578     & \multicolumn{1}{c|}{440}  & 1161    & 2446        & 394     \\
6                                              & 14035  & 2203     & \multicolumn{1}{c|}{8531}  & 0        & 0           & 0       & 3531  & 513      & \multicolumn{1}{c|}{2150} & 0       & 0           & 0       \\ \hline
Total                                          & 137036 & 29917    & \multicolumn{1}{c|}{20561} & 25811    & 17594       & 2723    & 34520 & 7522     & \multicolumn{1}{c|}{5143} & 6510    & 4493        & 719     \\ \bottomrule
\end{tabular}%
}
\end{table}

\section{Benchmarks}

In this section, we provide benchmarks of three typical tasks on our proposed DOLPHINS dataset: 2D object detection, 3D object detection, and multi-view collaborative perception. For each task, we implement several classical algorithms.

\subsection{Metrics}

We firstly aggregate the training datasets of six scenarios altogether. The composed dataset will contain various background characteristics and occlusion relationships, which helps the model to have better generalization ability. The training dataset is split for training and validation at the ratio of 5:3, then the performance of each detector is examined on the test dataset. Similar to KITTI~\cite{Geiger2012CVPR}, we use Average Precision (AP) at Intersection-over-Union (IoU) threshold of 0.7 and 0.5 to illustrate the goodness of detectors on cars, as well as IoU of 0.5 and 0.25 for the pedestrians since the pedestrians are much smaller than cars. The degree of difficulty is cumulative in the test, that is, the ground truths of easy objects are also considered in moderate and hard tests. 

\subsection{Experiment details}

We use MMDetection~\cite{mmdetection} and MMDetection3D~\cite{mmdet3d2020} to construct the training and test pipeline. As for 2D object detection tasks, we finetune the COCO~\cite{coco} pre-trained models on our dataset. We also provide the GPU memory consumption and the inference speed to illustrate the differences between different methods, where the experiment is set with a batch size equal to 1. All the experiments are performed on 8 RTX 3090 GPUs.

\subsection{2D object detection}

As mentioned in Sec. \ref{sec:related}, there are four typical detection paradigms: anchor-based two-stage detectors, anchor-based one-stage detectors, anchor-free one-stage detectors, and the vision transformer. In this part, we select Faster R-CNN~\cite{ren2015faster} as the representative of anchor-based two-stage detectors, YOLOv3~\cite{redmon2018yolov3} for the anchor-based one-stage detectors, YOLOX~\cite{yolox2021} and TOOD~\cite{feng2021tood} for the anchor-free one-stage detectors, and DETR~\cite{detr} for the vision transformer. Specifically, we set the backbone network of Faster-RCNN and YOLOv3 to be Resnet-50, so that the size of these networks is close to each other. The experiment results are illustrated in Table~\ref{tab:2D-speed} and Table~\ref{tab:2D-result}. It shows that all the detectors can have good knowledge of different scenarios. However, the modern anchor-free detectors can significantly speed up the entire inference procedure without loss of precision. One abnormal result is the surprising rise of AP in hard and moderate tasks compared with easy tasks, especially in the pedestrian detection. A reasonable explanation is the large proportion of moderate and hard objects due to the characteristics of different scenarios. For example, in Scenario 6, which is a mountain road, nearly half of the objects are severely occluded due to the undulating planes. Thus, the detectors tend to propose much more candidate objects to match those hard objects, which leads to low AP in easy tasks because of the false positives. It is proved by the high recall scores in easy tasks. The same thing happens in pedestrian detection, where the pedestrians are smaller and thus more likely to be hard ones. However, low AP is not equal to poor performance. On the contrary, meeting the ultra-reliability demands of self-driving, a higher recall rate is much more meaningful than the AP, which can alert the vehicles to the potential dangers in blind zones. More detailed analysis can be found in the supplementary material.

\vspace{-0.8cm}
\begin{table}
\centering
\caption{2D object detection analysis on speed and cost}
\label{tab:2D-speed}
\resizebox{0.5\textwidth}{!}{%
\begin{tabular}{@{}ccc@{}}
\toprule
Method      & Inference speed (fps) & Memory usage (MB) \\ \midrule
Faster-RCNN &  35.6                  &  2513                 \\
YOLOv3      &  50.7                  &  2285                 \\
YOLOX-S     &  58.1                   &  2001                 \\
YOLOX-L     &  36.5                  &  2233                 \\
TOOD        &  26.8                  &  2247                 \\
DETR        &  26.3                  &  2419                 \\ \bottomrule
\end{tabular}%
}
\end{table}
\vspace{-1cm}

\subsection{3D object detection}

As for the 3D object detection tasks, different modals of sensors lead to different detector architectures. We choose SECOND~\cite{yan2018second}, PointPillars~\cite{lang2019pointpillars}, and PV-RCNN~\cite{shi2020pv} as SOTA LiDAR-based methods in this part. What's more, the multi-modal detectors can combine the segmentation information from images and the depth information from LiDARs, which is an advantage to the detection of small objects which reflect few points, e.g. pedestrians. We also test MVX-Net~\cite{sindagi2019mvx} on our multi-modality dataset. The experiment results are illustrated in Table~\ref{tab:3D-speed} and Table~\ref{tab:3D-result}. The results show that Scenarios 3 and 6 are the corner cases where the AP is significantly lower than in other scenarios. Due to the steep ramp, the LiDAR on the ego vehicle is hard to detect the opposite vehicles and pedestrians, which is the fundamental defect of stand-alone intelligence. What's more, PV-RCNN~\cite{shi2020pv} gains significantly better performance at the cost of taking nearly four times as long as SECOND~\cite{yan2018second}. MVX-Net~\cite{sindagi2019mvx} is inferior to those pure LiDAR-based methods, but it achieves surprisingly performance in pedestrians, which means the rich segmentation of information from images is profit to the detection of small objects.

\subsection{Multi-view collaborative perception}

Based on the information to exchange, collaborative perception can be categorized into three levels: raw-level (early fusion), feature-level (middle fusion), and object-level (late fusion). Due to the 3D information provided by point clouds, LiDAR-based data fusion and object detection have been widely discussed. We realize a raw-level fusion algorithm based on DOLPHINS LiDAR data through the superposition of point clouds from different perspectives. However, not all the LiDAR-based 3D detection algorithms can be adapted to raw-level fusion schemes. Since many detectors use voxels to represent the point clouds of a district, the height of voxels is limited to reduce the computation complexity. The limitation will not be violated when the cooperators are on the same horizontal plane, as in \cite{chen2019cooper} and \cite{opv2v}. However, when the data are from RSUs or from vehicles on a mountain road (as in Scenario 3 and 6 in Fig.~\ref{fig:scenario}), the height of the aggregated point clouds will be too large to tackle through traditional voxel processing. In our experiment settings, PointPillars~\cite{lang2019pointpillars} is the only algorithm to be compatible with the raw-level fusion scheme.

\begin{table}
\centering
\caption{2D object detection results on DOLPHINS}
\label{tab:2D-result}
\resizebox{\textwidth}{!}{%
\begin{tabular}{c|c|ccc|ccc|ccc|ccc}
\toprule
\multirow{3}{*}{Scenario} & \multirow{3}{*}{Method} & \multicolumn{3}{c|}{\multirow{2}{*}{\begin{tabular}[c]{@{}c@{}}Car\\ AP@IoU=0.7\end{tabular}}} & \multicolumn{3}{c|}{\multirow{2}{*}{\begin{tabular}[c]{@{}c@{}}Car\\ AP@IoU=0.5\end{tabular}}} & \multicolumn{3}{c|}{\multirow{2}{*}{\begin{tabular}[c]{@{}c@{}}Pedestrian\\ AP@IoU=0.5\end{tabular}}} & \multicolumn{3}{c}{\multirow{2}{*}{\begin{tabular}[c]{@{}c@{}}Pedestrian\\ AP@IoU=0.25\end{tabular}}} \\
                          &                         & \multicolumn{3}{c|}{}                                                                          & \multicolumn{3}{c|}{}                                                                          & \multicolumn{3}{c|}{}                                                                                 & \multicolumn{3}{c}{}                                                                                  \\ \cline{3-14} 
                          &                         & Easy                          & Moderate                        & Hard                         & Easy                          & Moderate                        & Hard                         & Easy                            & Moderate                           & Hard                           & Easy                            & Moderate                           & Hard                           \\ \midrule
\multirow{6}{*}{1}        & Faster R-CNN            & 89.18                         & 80.73                           & 80.62                        & 90.07                         & 89.72                           & 88.76                        & 87.12                           & 90.77                              & 90.82                          & 87.12                           & 90.79                              & 90.88                          \\
                          & YOLOv3                  & 84.76                         & 79.36                           & 79.19                        & 86.93                         & 89.90                           & 89.79                        & 85.22                           & 90.41                              & 90.54                          & 85.40                           & 90.66                              & 90.79                          \\
                          & YOLOX-S                 & 76.21                         & 69.18                           & 68.31                        & 87.57                         & 84.79                           & 82.44                        & 87.19                           & 90.21                              & 89.54                          & 87.46                           & 90.63                              & 90.40                          \\
                          & YOLOX-L                 & 84.72                         & 79.58                           & 78.79                        & 88.95                         & 88.90                           & 87.42                        & 87.76                           & 90.66                              & 90.70                          & 87.82                           & 90.73                              & 90.81                          \\
                          & TOOD                    & 88.30                         & 80.04                           & 79.97                        & 89.76                         & 90.07                           & 89.71                        & 89.02                           & 90.83                              & 90.87                          & 89.03                           & 90.83                              & 90.88                          \\
                          & DETR                    & 82.29                         & 77.76                           & 75.61                        & 89.40                         & 88.43                           & 86.79                        & 87.16                           & 89.44                              & 89.19                          & 87.77                           & 90.17                              & 90.14                          \\ \hline
\multirow{6}{*}{2}        & Faster R-CNN            & 90.50                         & 90.31                           & 90.19                        & 90.82                         & 90.81                           & 90.80                        & 87.36                           & 90.49                              & 90.18                          & 87.36                           & 90.66                              & 90.60                          \\
                          & YOLOv3                  & 89.33                         & 89.46                           & 89.48                        & 90.17                         & 90.27                           & 90.47                        & 75.96                           & 90.05                              & 90.11                          & 75.96                           & 90.47                              & 90.59                          \\
                          & YOLOX-S                 & 86.46                         & 79.98                           & 79.53                        & 90.20                         & 89.85                           & 89.54                        & 86.99                           & 86.70                              & 80.76                          & 87.24                           & 89.66                              & 87.92                          \\
                          & YOLOX-L                 & 89.48                         & 86.87                           & 80.85                        & 90.61                         & 90.47                           & 90.25                        & 86.29                           & 89.51                              & 87.65                          & 86.29                           & 89.98                              & 89.72                          \\
                          & TOOD                    & 90.39                         & 90.20                           & 90.10                        & 90.85                         & 90.84                           & 90.82                        & 88.12                           & 90.75                              & 90.53                          & 93.67                           & 90.80                              & 90.76                          \\
                          & DETR                    & 89.90                         & 88.52                           & 87.30                        & 90.60                         & 90.53                           & 90.47                        & 86.80                           & 89.63                              & 87.12                          & 89.83                           & 90.51                              & 90.12                          \\ \hline
\multirow{6}{*}{3}        & Faster R-CNN            & 85.23                         & 88.55                           & 80.25                        & 86.54                         & 90.18                           & 81.51                        & 75.89                           & 89.77                              & 90.67                          & 75.89                           & 89.80                              & 90.76                          \\
                          & YOLOv3                  & 84.64                         & 78.65                           & 71.03                        & 87.46                         & 88.97                           & 80.94                        & 44.31                           & 88.77                              & 89.23                          & 44.63                           & 89.80                              & 90.74                          \\
                          & YOLOX-S                 & 85.95                         & 76.67                           & 66.72                        & 89.42                         & 87.02                           & 77.75                        & 82.54                           & 87.99                              & 81.18                          & 86.54                           & 89.91                              & 85.83                          \\
                          & YOLOX-L                 & 88.55                         & 80.41                           & 71.25                        & 89.64                         & 89.70                           & 80.39                        & 86.55                           & 90.39                              & 89.65                          & 89.02                           & 90.43                              & 90.37                          \\
                          & TOOD                    & 88.25                         & 80.47                           & 71.50                        & 89.62                         & 90.30                           & 81.26                        & 85.60                           & 90.15                              & 90.69                          & 85.60                           & 90.18                              & 90.83                          \\
                          & DETR                    & 87.94                         & 85.00                           & 79.18                        & 89.17                         & 88.71                           & 86.74                        & 85.73                           & 87.49                              & 85.87                          & 85.73                           & 88.69                              & 87.56                          \\ \hline
\multirow{6}{*}{4}        & Faster R-CNN            & 89.33                         & 81.21                           & 81.09                        & 89.40                         & 89.68                           & 88.70                        & N/A                             & N/A                                & N/A                            & N/A                             & N/A                                & N/A                            \\
                          & YOLOv3                  & 82.43                         & 78.22                           & 77.66                        & 83.95                         & 89.21                           & 89.50                        & N/A                             & N/A                                & N/A                            & N/A                             & N/A                                & N/A                            \\
                          & YOLOX-S                 & 88.97                         & 76.37                           & 70.29                        & 89.65                         & 86.24                           & 84.43                        & N/A                             & N/A                                & N/A                            & N/A                             & N/A                                & N/A                            \\
                          & YOLOX-L                 & 89.96                         & 80.81                           & 79.58                        & 90.07                         & 89.20                           & 88.25                        & N/A                             & N/A                                & N/A                            & N/A                             & N/A                                & N/A                            \\
                          & TOOD                    & 90.08                         & 81.19                           & 80.22                        & 90.22                         & 90.04                           & 81.63                        & N/A                             & N/A                                & N/A                            & N/A                             & N/A                                & N/A                            \\
                          & DETR                    & 88.17                         & 76.00                           & 74.13                        & 89.44                         & 86.50                           & 86.00                        & N/A                             & N/A                                & N/A                            & N/A                             & N/A                                & N/A                            \\ \hline
\multirow{6}{*}{5}        & Faster R-CNN            & 89.70                         & 80.86                           & 81.04                        & 90.25                         & 89.34                           & 88.82                        & 68.07                           & 80.10                              & 81.06                          & 71.49                           & 80.17                              & 81.41                          \\
                          & YOLOv3                  & 83.57                         & 78.58                           & 78.80                        & 85.45                         & 89.64                           & 89.77                        & 38.03                           & 87.20                              & 86.48                          & 40.79                           & 89.25                              & 90.52                          \\
                          & YOLOX-S                 & 79.04                         & 75.71                           & 70.72                        & 88.53                         & 86.92                           & 85.07                        & 65.48                           & 79.95                              & 77.82                          & 65.54                           & 85.47                              & 79.99                          \\
                          & YOLOX-L                 & 87.88                         & 80.14                           & 80.18                        & 89.20                         & 89.53                           & 88.75                        & 64.05                           & 87.73                              & 81.18                          & 64.05                           & 88.61                              & 87.24                          \\
                          & TOOD                    & 89.12                         & 85.12                           & 80.73                        & 89.87                         & 89.92                           & 89.96                        & 77.25                           & 88.18                              & 81.41                          & 78.71                           & 89.16                              & 88.43                          \\
                          & DETR                    & 84.26                         & 78.03                           & 75.95                        & 89.58                         & 88.35                           & 87.00                        & 74.59                           & 81.21                              & 77.55                          & 77.04                           & 86.34                              & 83.82                          \\ \hline
\multirow{6}{*}{6}        & Faster R-CNN            & 77.97                         & 79.60                           & 90.10                        & 78.09                         & 79.74                           & 90.55                        & N/A                             & N/A                                & N/A                            & N/A                             & N/A                                & N/A                            \\
                          & YOLOv3                  & 79.67                         & 79.68                           & 88.85                        & 80.00                         & 80.12                           & 90.35                        & N/A                             & N/A                                & N/A                            & N/A                             & N/A                                & N/A                            \\
                          & YOLOX-S                 & 81.63                         & 79.53                           & 79.73                        & 82.34                         & 81.28                           & 88.61                        & N/A                             & N/A                                & N/A                            & N/A                             & N/A                                & N/A                            \\
                          & YOLOX-L                 & 76.59                         & 76.87                           & 87.35                        & 76.85                         & 77.32                           & 90.07                        & N/A                             & N/A                                & N/A                            & N/A                             & N/A                                & N/A                            \\
                          & TOOD                    & 82.54                         & 81.62                           & 90.63                        & 82.65                         & 81.77                           & 89.93                        & N/A                             & N/A                                & N/A                            & N/A                             & N/A                                & N/A                            \\
                          & DETR                    & 85.09                         & 83.27                           & 81.81                        & 85.47                         & 84.03                           & 89.89                        & N/A                             & N/A                                & N/A                            & N/A                             & N/A                                & N/A                            \\ \bottomrule
\end{tabular}%
}
\end{table}

\begin{table}
\centering
\caption{3D object detection analysis on speed and cost}
\label{tab:3D-speed}
\resizebox{0.5\textwidth}{!}{%
\begin{tabular}{@{}ccc@{}}
\toprule
Method      & Inference speed (fps) & Memory usage (MB) \\ \midrule
SECOND &            45.7          &         2433          \\
PointPillars      &     36.8                 &      3483             \\
PV-RCNN       &             13.1         &          2899         \\
MVX-Net        &        11.0              &      11321             \\ \bottomrule
\end{tabular}%
}
\end{table}

What's more, we also extend the MVX-Net to the collaborative autonomous driving scenarios. With the help of the point clouds from the LiDARs on the RSUs, which locations are usually much higher, the ego vehicle can have a wider view with fewer occlusions. In addition, a single LiDAR on the RSU could free all the nearby autonomous vehicles from the necessity of equipping expensive LiDARs by sharing its point clouds through the V2I network, which brings great benefits to the realization of Level-5 autonomous driving. In this work, we use the point clouds from the RSU (or the aux vehicle 1 in Scenario 5) instead of the ego vehicle by transforming the coordinates. 

Table~\ref{tab:cp-result} illustrates the multi-view collaborative perception on PointPillars~\cite{lang2019pointpillars} and MVX-Net~\cite{sindagi2019mvx}, and the corresponding AP difference compared with stand-alone detection. It turns out that as for the superposition of raw point clouds, the ego vehicle can gain plentiful benefits from the richer information directly from another perspective. Under those circumstances with severe occlusions such as Scenario 3 and 6 and for those hard objects, the cooperative perception-based PointPillars~\cite{lang2019pointpillars} achieves up to 38.42\% increment in AP. However, the extra noise also infects the detection of small objects. On the other hand, as for the MVX-Net with the local camera and RSU LiDAR, the performance is nearly the same as the one with stand-alone sensors. It shows the opportunity to enable high-level autonomous driving on cheap vehicles, which are not equipped with LiDARs, through the sensors on infrastructures.

\begin{table}
\centering
\caption{3D object detection results on DOLPHINS}
\label{tab:3D-result}
\resizebox{\textwidth}{!}{%
\begin{tabular}{c|c|ccc|ccc|ccc|ccc}
\toprule
\multirow{3}{*}{Scenario} & \multirow{3}{*}{Method} & \multicolumn{3}{c|}{\multirow{2}{*}{\begin{tabular}[c]{@{}c@{}}Car\\ AP@IoU=0.7\end{tabular}}} & \multicolumn{3}{c|}{\multirow{2}{*}{\begin{tabular}[c]{@{}c@{}}Car\\ AP@IoU=0.5\end{tabular}}} & \multicolumn{3}{c|}{\multirow{2}{*}{\begin{tabular}[c]{@{}c@{}}Pedestrian\\ AP@IoU=0.5\end{tabular}}} & \multicolumn{3}{c}{\multirow{2}{*}{\begin{tabular}[c]{@{}c@{}}Pedestrian\\ AP@IoU=0.25\end{tabular}}} \\
                          &                         & \multicolumn{3}{c|}{}                                                                          & \multicolumn{3}{c|}{}                                                                          & \multicolumn{3}{c|}{}                                                                                 & \multicolumn{3}{c}{}                                                                                  \\ \cline{3-14} 
                          &                         & Easy                          & Moderate                        & Hard                         & Easy                          & Moderate                        & Hard                         & Easy                            & Moderate                           & Hard                           & Easy                            & Moderate                           & Hard                           \\ \midrule
\multirow{4}{*}{1}        & SECOND                  & 95.65                         & 90.37                           & 87.36                        & 98.79                         & 96.05                           & 92.97                        & 74.17                           & 70.08                              & 68.19                          & 96.78                           & 95.86                              & 93.47                          \\
                          & PointPillar             & 96.63                         & 92.09                           & 88.83                        & 98.55                         & 96.19                           & 93.13                        & 70.12                           & 65.41                              & 63.56                          & 95.57                           & 94.38                              & 91.98                          \\
                          & PV-RCNN                 & 98.14                         & 93.87                           & 90.69                        & 98.90                         & 96.18                           & 92.98                        & 83.67                           & 80.36                              & 78.37                          & 97.31                           & 96.50                              & 94.09                          \\
                          & MVX-Net                 & 89.25                         & 84.3                            & 83.99                        & 89.62                         & 89.51                           & 87.02                        & 91.36                           & 88.93                              & 86.49                          & 99.61                           & 99.55                              & 97.06                          \\ \hline
\multirow{4}{*}{2}        & SECOND                  & 96.33                         & 91.32                           & 88.77                        & 98.48                         & 97.50                           & 95.56                        & 66.71                           & 59.35                              & 57.27                          & 95.07                           & 91.00                              & 87.94                          \\
                          & PointPillar             & 97.39                         & 93.06                           & 91.70                        & 98.23                         & 97.33                           & 95.48                        & 59.65                           & 53.33                              & 51.35                          & 92.52                           & 87.58                              & 84.49                          \\
                          & PV-RCNN                 & 98.57                         & 94.76                           & 91.15                        & 99.12                         & 97.44                           & 94.33                        & 78.74                           & 71.59                              & 68.94                          & 97.37                           & 93.91                              & 90.60                          \\
                          & MVX-Net                 & 93.77                         & 91.04                           & 88.63                        & 96.33                         & 95.91                           & 93.72                        & 89.12                           & 82.22                              & 79.78                          & 99.37                           & 98.72                              & 94.19                          \\ \hline
\multirow{4}{*}{3}        & SECOND                  & 80.30                         & 67.16                           & 54.96                        & 85.51                         & 75.08                           & 64.29                        & 49.87                           & 29.31                              & 27.05                          & 92.73                           & 60.26                              & 55.88                          \\
                          & PointPillar             & 78.94                         & 68.29                           & 56.78                        & 85.67                         & 75.75                           & 66.75                        & 37.49                           & 22.04                              & 20.00                          & 82.64                           & 52.55                              & 48.25                          \\
                          & PV-RCNN                 & 85.93                         & 73.85                           & 60.59                        & 87.54                         & 77.18                           & 64.30                        & 63.95                           & 37.46                              & 34.53                          & 90.93                           & 58.96                              & 54.32                          \\
                          & MVX-Net                 & 68.96                         & 58.47                           & 48.73                        & 71.80                         & 61.67                           & 56.25                        & 71.96                           & 41.36                              & 37.10                          & 93.99                           & 56.62                              & 53.85                          \\ \hline
\multirow{4}{*}{4}        & SECOND                  & 97.81                         & 92.11                           & 84.47                        & 99.33                         & 97.35                           & 90.50                        & N/A                             & N/A                                & N/A                            & N/A                             & N/A                                & N/A                            \\
                          & PointPillar             & 98.07                         & 94.00                           & 86.52                        & 98.79                         & 97.57                           & 91.01                        & N/A                             & N/A                                & N/A                            & N/A                             & N/A                                & N/A                            \\
                          & PV-RCNN                 & 99.37                         & 95.54                           & 87.70                        & 99.50                         & 97.78                           & 89.98                        & N/A                             & N/A                                & N/A                            & N/A                             & N/A                                & N/A                            \\
                          & MVX-Net                 & 91.76                         & 86.39                           & 83.73                        & 91.96                         & 89.19                           & 86.55                        & N/A                             & N/A                                & N/A                            & N/A                             & N/A                                & N/A                            \\ \hline
\multirow{4}{*}{5}        & SECOND                  & 96.49                         & 91.41                           & 87.44                        & 98.68                         & 96.36                           & 92.71                        & 75.33                           & 65.03                              & 58.90                          & 97.37                           & 94.29                              & 87.52                          \\
                          & PointPillar             & 97.45                         & 92.92                           & 89.30                        & 98.81                         & 96.52                           & 92.94                        & 71.36                           & 62.07                              & 56.27                          & 97.32                           & 93.59                              & 85.70                          \\
                          & PV-RCNN                 & 98.57                         & 94.39                           & 90.49                        & 99.23                         & 97.21                           & 93.15                        & 90.47                           & 79.07                              & 71.64                          & 98.94                           & 95.45                              & 87.29                          \\
                          & MVX-Net                 & 91.69                         & 86.67                           & 84.17                        & 94.45                         & 91.94                           & 89.39                        & 84.87                           & 75.83                              & 68.91                          & 99.52                           & 99.36                              & 91.81                          \\ \hline
\multirow{4}{*}{6}        & SECOND                  & 90.53                         & 82.60                           & 56.05                        & 97.54                         & 82.11                           & 68.15                        & N/A                             & N/A                                & N/A                            & N/A                             & N/A                                & N/A                            \\
                          & PointPillar             & 89.31                         & 82.30                           & 57.32                        & 97.44                         & 92.66                           & 70.62                        & N/A                             & N/A                                & N/A                            & N/A                             & N/A                                & N/A                            \\
                          & PV-RCNN                 & 95.95                         & 89.29                           & 62.05                        & 98.23                         & 93.95                           & 69.28                        & N/A                             & N/A                                & N/A                            & N/A                             & N/A                                & N/A                            \\
                          & MVX-Net                 & 87.53                         & 75.29                           & 52.60                        & 90.76                         & 80.55                           & 57.97                        & N/A                             & N/A                                & N/A                            & N/A                             & N/A                                & N/A                            \\ \bottomrule
\end{tabular}%
}
\end{table}

\begin{table}
\centering
\caption{Multi-view collaborative perception results on DOLPHINS}
\label{tab:cp-result}
\resizebox{\textwidth}{!}{%
\begin{tabular}{c|c|ccc|ccc|ccc|ccc}
\hline
\multirow{3}{*}{Scenario} & \multirow{3}{*}{Method} & \multicolumn{3}{c|}{\multirow{2}{*}{\begin{tabular}[c]{@{}c@{}}Car\\ AP@IoU=0.7\end{tabular}}} & \multicolumn{3}{c|}{\multirow{2}{*}{\begin{tabular}[c]{@{}c@{}}Car\\ AP@IoU=0.5\end{tabular}}} & \multicolumn{3}{c|}{\multirow{2}{*}{\begin{tabular}[c]{@{}c@{}}Pedestrian\\ AP@IoU=0.5\end{tabular}}} & \multicolumn{3}{c}{\multirow{2}{*}{\begin{tabular}[c]{@{}c@{}}Pedestrian\\ AP@IoU=0.25\end{tabular}}} \\
                          &                         & \multicolumn{3}{c|}{}                                                                          & \multicolumn{3}{c|}{}                                                                          & \multicolumn{3}{c|}{}                                                                                 & \multicolumn{3}{c}{}                                                                                  \\ \cline{3-14} 
                          &                         & Easy                          & Moderate                        & Hard                         & Easy                          & Moderate                        & Hard                         & Easy                             & Moderate                          & Hard                           & Easy                             & Moderate                          & Hard                           \\ \hline
\multirow{4}{*}{1}        & PointPillar CP          & 97.13                         & 95.19                           & 94.57                        & 97.63                         & 96.40                           & 95.89                        & 72.73                            & 70.60                             & 70.23                          & 94.01                            & 93.69                             & 93.31                          \\
                          & Difference              & 0.52\%                         & 3.37\%                           & 6.46\%                        & -0.93\%                        & 0.22\%                           & 2.96\%                        & 3.72\%                            & 7.93\%                             & 10.49\%                         & -1.63\%                           & -0.73\%                            & 1.45\%                          \\
                          & MVX-Net CP              & 89.34                         & 84.30                           & 84.02                        & 89.64                         & 89.49                           & 87.00                        & 90.77                            & 86.08                             & 85.92                          & 99.61                            & 99.53                             & 97.04                          \\
                          & Difference              & 0.10\%                         & 0.00\%                           & 0.04\%                        & 0.02\%                         & -0.02\%                          & -0.02\%                       & -0.65\%                           & -3.20\%                            & -0.66\%                         & 0.00\%                            & -0.02\%                            & -0.02\%                         \\ \hline
\multirow{4}{*}{2}        & PointPillar CP          & 97.97                         & 97.03                           & 96.42                        & 98.52                         & 97.89                           & 97.55                        & 57.51                            & 51.89                             & 50.92                          & 91.94                            & 88.52                             & 86.54                          \\
                          & Difference              & 0.60\%                         & 4.27\%                           & 5.15\%                        & 0.30\%                         & 0.58\%                           & 2.17\%                        & -3.59\%                           & -2.70\%                            & -0.84\%                         & -0.63\%                           & 1.07\%                             & 2.43\%                          \\
                          & MVX-Net CP              & 93.75                         & 90.99                           & 88.47                        & 96.41                         & 96.19                           & 93.81                        & 88.52                            & 81.72                             & 79.25                          & 99.29                            & 98.76                             & 94.13                          \\
                          & Difference              & -0.02\%                        & -0.05\%                          & -0.18\%                       & 0.08\%                         & 0.29\%                           & 0.10\%                        & -0.67\%                           & -0.61\%                            & -0.66\%                         & -0.08\%                           & 0.04\%                             & -0.06\%                         \\ \hline
\multirow{4}{*}{3}        & PointPillar CP          & 81.48                         & 72.73                           & 66.30                        & 86.88                         & 77.69                           & 72.33                        & 32.62                            & 19.89                             & 18.47                          & 73.24                            & 48.12                             & 44.80                          \\
                          & Difference              & 3.22\%                         & 6.50\%                           & 16.77\%                       & 1.41\%                         & 2.56\%                           & 8.36\%                        & -12.99\%                          & -9.75\%                            & -7.65\%                         & -11.37\%                          & -8.43\%                            & -7.15\%                         \\
                          & MVX-Net CP              & 69.04                         & 58.48                           & 48.77                        & 71.61                         & 61.60                           & 54.18                        & 70.82                            & 39.76                             & 37.43                          & 91.87                            & 56.60                             & 51.76                          \\
                          & Difference              & 0.12\%                         & 0.02\%                           & 0.08\%                        & -0.26\%                        & -0.11\%                          & -3.68\%                       & -1.58\%                           & -3.87\%                            & 0.89\%                          & -2.26\%                           & -0.04\%                            & -3.88\%                         \\ \hline
\multirow{4}{*}{4}        & PointPillar CP          & 97.60                         & 96.22                           & 94.40                        & 97.93                         & 97.00                           & 95.74                        & N/A                              & N/A                               & N/A                            & N/A                              & N/A                               & N/A                            \\
                          & Difference              & -0.48\%                        & 2.36\%                           & 9.11\%                        & -0.87\%                        & -0.58\%                          & 5.20\%                        & N/A                              & N/A                               & N/A                            & N/A                              & N/A                               & N/A                            \\
                          & MVX-Net CP              & 91.70                         & 86.38                           & 83.69                        & 91.92                         & 89.22                           & 86.58                        & N/A                              & N/A                               & N/A                            & N/A                              & N/A                               & N/A                            \\
                          & Difference              & -0.07\%                        & -0.01\%                          & -0.05\%                       & -0.04\%                        & 0.03\%                           & 0.03\%                        & N/A                              & N/A                               & N/A                            & N/A                              & N/A                               & N/A                            \\ \hline
\multirow{4}{*}{5}        & PointPillar CP          & 96.38                         & 94.24                           & 92.77                        & 96.87                         & 95.82                           & 94.38                        & 65.50                            & 61.53                             & 58.69                          & 93.00                            & 91.62                             & 87.82                          \\
                          & Difference              & -1.10\%                        & 1.42\%                           & 3.89\%                        & -1.96\%                        & -0.73\%                          & 1.55\%                        & -8.21\%                           & -0.87\%                            & 4.30\%                          & -4.44\%                           & -2.10\%                            & 2.47\%                          \\
                          & MVX-Net CP              & 91.63                         & 86.57                           & 84.06                        & 94.49                         & 91.94                           & 89.40                        & 82.56                            & 72.02                             & 67.19                          & 99.78                            & 96.93                             & 89.44                          \\
                          & Difference              & -0.07\%                        & -0.12\%                          & -0.13\%                       & 0.04\%                         & 0.00\%                           & 0.01\%                        & -2.72\%                           & -5.02\%                            & -2.50\%                         & 0.26\%                            & -2.45\%                            & -2.58\%                         \\ \hline
\multirow{4}{*}{6}        & PointPillar CP          & 94.46                         & 91.43                           & 79.34                        & 97.77                         & 96.63                           & 87.17                        & N/A                              & N/A                               & N/A                            & N/A                              & N/A                               & N/A                            \\
                          & Difference              & 5.77\%                         & 11.09\%                          & 38.42\%                       & 0.34\%                         & 4.28\%                           & 23.44\%                       & N/A                              & N/A                               & N/A                            & N/A                              & N/A                               & N/A                            \\
                          & MVX-Net CP              & 87.10                         & 74.85                           & 52.34                        & 90.45                         & 80.43                           & 59.52                        & N/A                              & N/A                               & N/A                            & N/A                              & N/A                               & N/A                            \\
                          & Difference              & -0.49\%                        & -0.58\%                          & -0.49\%                       & -0.34\%                        & -0.15\%                          & 2.67\%                        & N/A                              & N/A                               & N/A                            & N/A                              & N/A                               & N/A                            \\ \hline
\end{tabular}%
}
\end{table}

\section{Conclusions}

In this paper, we present a new large-scale various-scenario multi-view multi-modality autonomous driving dataset, DOLPHINS, to facilitate the research on collaborative perception-enabled connected autonomous driving. All the data are temporally-aligned and generated from three viewpoints, including both vehicles and RSUs, in six typical driving scenarios, along with the annotations, calibrations, and the geo-positions. What's more, we benchmark several SOTA algorithms on traditional 2D/3D object detection and brand-new collaborative perception tasks. The experiment results suggest that not only the extra data from V2X communication can eliminate the occlusions, but also the RSUs at appropriate locations can provide equivalent point clouds to the nearby vehicles, which can greatly reduce the prime cost of self-driving cars. In the future, we are going to further extend the number of infrastructures and aux vehicles, and construct more realistic maps of the downtown.



\bibliographystyle{splncs}
\bibliography{egbib}

\end{document}